\documentclass[shortpaper]{clv2} 

\usepackage{graphicx}

\usepackage{linguex}
\usepackage{longtable}
\usepackage{subcaption}

\usepackage[table]{xcolor}
\usepackage{multirow}
\usepackage{hhline}
\usepackage{color,array}
 
\definecolor{sfbgreen}{RGB}{61,144,28}
\definecolor{darkgreen}{RGB}{0,102,0}
\definecolor{heat1}{RGB}{80,120,50} 
\definecolor{heat2}{RGB}{101,133,70} 
\definecolor{heat3}{RGB}{125,160,98} 
\definecolor{heat4}{RGB}{130,170,100} 
\definecolor{heat5}{RGB}{139,185,110} 
\definecolor{heat6}{RGB}{154,199,124} 
\definecolor{heat7}{RGB}{169,202,145} 
\definecolor{heat8}{RGB}{180,210,147} 
\definecolor{heat9}{RGB}{187,216,156} 
\definecolor{heat10}{RGB}{201,229,170} 

\definecolor{red1}{RGB}{230,70,55} 
\definecolor{red4}{RGB}{240,100,60} 
\definecolor{red5}{RGB}{245,130,71} 
\definecolor{red6}{RGB}{248,177,82} 
\definecolor{red7}{RGB}{250,217,120} 
\definecolor{red8}{RGB}{252,253,209} 

\usepackage{hyperref}

\issue{X}{X}{XXXX}

\dochead{How compatible are our discourse annotation frameworks? \\Insights from mapping RST-DT and PDTB annotations}

\runningtitle{How compatible are our discourse annotation frameworks?}

\runningauthor{Demberg, Asr, and Scholman}

\begin{document}

\newcommand{\sid}[1]{{\begin{sideways} #1 \end{sideways}}}
\newcommand{\expect}[1]{{\underline{{\bf #1}}}}

\newcommand{\argo}[1]{{\em #1}} 
\newcommand{\argt}[1]{{\bf #1}}  
\newcommand{\conn}[1]{{\em #1}}
\newcommand{\connpdtb}[1]{{\underline{#1}}}
\newcommand{\rel}[1]{{\sc #1}}
\newcommand{\ccrvalue}[1]{{\texttt{#1}}}

\author{Vera Demberg\thanks{Department of Computer Science, Saarland University, Saarbr\"ucken, 66123, Germany. \newline E-mail: vera@coli.uni-saarland.de}\textsuperscript{,$\dagger$}}
\affil{Saarland University}

\author{Fatemeh Torabi Asr\thanks{Department of Linguistics, Simon Fraser University, Burnaby, BC V5A 1S6, Canada. \newline E-mail: ftorabia@sfu.ca}}
\affil{Simon Fraser University}

\author{Merel Scholman\thanks{Department of Language Science and Technology, Saarland University, Saarbr\"ucken, 66123, Germany. \newline E-mail: m.c.j.scholman@coli.uni-saarland.de}}
\affil{Saarland University}

\maketitle

\begin{abstract}

%
Discourse-annotated corpora are an important resource for the community, but they are often annotated according to different frameworks. This makes comparison of the annotations difficult, thereby also preventing researchers from searching the corpora in a unified way, or using all annotated data jointly to train computational systems.
Several theoretical proposals have recently been made for mapping the relational labels of different frameworks to each other, but these proposals have so far not been validated against existing annotations.
The two largest discourse relation annotated resources, the Penn Discourse Treebank and the Rhetorical Structure Theory Discourse Treebank, have however been annotated on the same text, allowing for a direct comparison of the annotation layers. We propose a method for automatically aligning the discourse segments, and then evaluate existing mapping proposals by comparing the empirically observed against the proposed mappings.
Our analysis highlights the influence of segmentation on subsequent discourse relation labelling, and shows that while agreement between frameworks is reasonable for explicit relations, agreement on implicit relations is low. We identify several sources of systematic discrepancies between the two annotation schemes and  discuss consequences of these discrepancies for future annotation and for the training of automatic discourse relation labellers.
\end{abstract}

\section{Introduction}\label{sec:intro}

In recent years, we have seen an increase in attention to discourse processing, in terms of discourse relation labelling as a task in automatic language processing, as well as a feature for improving down-stream tasks such as machine translation \cite{meyer2012using, popescu2016manual}, question answering \cite{jansen2014discourse, sharp2015spinning} and sentiment analysis \cite{somasundaran2009supervised,zhou2011unsupervised,zirn2011fine}. Progress on this topic has been made possible through the large-scale annotation of text corpora with discourse relation labels,  most notably the Penn Discourse Treebank \cite[PDTB;][]{prasad2008} and the RST Treebank \cite[RST-DT;][]{carlson2003} for English. There are also new resources for other languages, as well as numerous annotation efforts currently under way \cite[see, for example,][]{oza2009,stede2014}.

As of yet there is however no consensus on a single discourse relation labelling scheme. 
Existing discourse frameworks share basic notions of what a coherence relation is, and many of them make relation sense distinctions that are based on similar underlying ideas, but frameworks differ in how they define discourse relational arguments, in terms of constraints on resulting discourse structure (e.g., whether it has to be a tree), and in whether they are ``lexically grounded'' (like PDTB-style annotations), semantically driven (like SDRT), or contain a combination of semantic and intentional relations (RST-DT).
This makes it difficult to study discourse relations across resources annotated according to different schemes, or across languages. In automatic discourse relation classifiers, it also limits the extent to which all available resources can be used effectively for training classifiers. This situation has long been recognized as a problem: in the early nineties, \namecite{hovy1995} taxonomized the more than 400 relations that have been proposed in different frameworks as a hierarchy of roughly 70 discourse relations.  

More recently, several large initiatives have addressed this issue as well: the COST initiative TextLink\footnote{\url{http://textlink.ii.metu.edu.tr}} is aimed at organizing the properties of discourse relations and encouraging the use of a single taxonomy for subsequent discourse annotation, as well as for searches in existing corpora. In this context, some concrete proposals have been made for how discourse relations may be mapped onto one another \cite{bunt2016,benamara2015,chiarcos2014,sanders2016}. 
\namecite{bunt2016} developed an ISO standard for coherence relations, in which they propose a new set of coherence relations that are central in many frameworks, and relate each of these to existing labels in other frameworks. The proposed set of labels can also be used for mapping labels to each other.
In a similar line of work, \namecite{benamara2015} proposed a unified set of 26 discourse relations based on distinctions made by several styles of the RST and SDRT frameworks. They compared discourse relation labels between the frameworks based on their definitions and their overall frequencies of occurrence, but they did not have any data available that was annotated according to both frameworks. They therefore did not evaluate whether the actual annotations of the two frameworks would correspond to one another.

\namecite{benamara2015}'s work highlights differences in granularity between frameworks by identifying certain labels that exist in one framework but do not have a corresponding label in the other framework. Such relations with no correspondence across taxonomies need more consideration when using an intermediate framework.
A possible solution for this mapping issue is to create an intermediate representation for mapping between frameworks, rather than creating a new framework. \namecite{chiarcos2014} was the first to attempt this: he developed an ontology to integrate RST-DT, PDTB and OntoNote annotations within a higher-level, more general framework. In this framework, the RST-DT and PDTB labels are assigned new labels with respect to the more general relation senses in both schemata. 
\namecite{sanders2016} created a different version of an intermediate representation. They worked out a mapping for discourse relations from various frameworks to a set of properties, such that each coherence relation can be described in terms of its properties or ``dimensions'' (such as ``causal'' vs.~``additive'', ``positive'' vs.~``negative'', and ``subjective'' vs.~``objective''). Through this intermediary representation in terms of properties, coherence relations can be mapped onto one another. 

As a community, we now find ourselves in a situation where several alternative proposals have been made for mapping coherence relation labels onto one another, but they have not been evaluated. These mappings were mainly proposed based on relation definitions, and we do not know whether alternative proposals for mapping relations are equivalent. Moreover, we do not know how well any of these proposals live up to the annotation of actual data. It is possible, for instance, that coherence relation labels should correspond to one another according to the annotation guidelines, but that differences in the operationalizations of frameworks (i.e., how annotators are asked to proceed for deciding on a relation label) lead to slightly different usages of these labels in actual annotations. Annotation manuals are also often (necessarily) incomplete -- they list clear cases or prototypical examples of certain types of coherence relations, but annotators learn through training and discussion how to deal with various kinds of difficult cases. The effect of such implicit knowledge on annotation practice may also lead to discrepancies between actual annotations compared to theoretically posited correspondences. When searching for a discourse phenomenon across several corpora, it may therefore not be sufficient to rely on the theoretically mapped labels. Instead, additional insights for which labels to consider or exclude can be gained from learning which labels correspond to one another empirically based on a large set of annotated instances.
This article therefore also aims to elucidate the extent to which aspects of operationalization or training may affect labelling decisions during annotation.

The PDTB 2.0 \cite{prasad2008,prasad2014reflections} and RST-DT  \cite{carlson2003} corpora represent an excellent opportunity for addressing these questions, as they have been annotated \textit{on the same text}. We will therefore focus on the PDTB to RST-DT label mappings that have been proposed by various researchers, and compare them against the correspondences between the annotations of actual instances found in the corpora. In the ideal case, we would expect to find (i) that the different mapping schemes are consistent with each other, i.e.~they propose the same set of equivalences between relations and (ii) that the proposed theoretical equivalences also hold for actual annotated data, i.e.~if a given text is annotated according to two different schemes, and there is a mapping between these schemes, the actual annotations should correspond to one another as specified by the theoretical mappings.

The current study thus extends previous work by mapping existing PDTB and RST-DT annotations onto one another, and comparing them to theoretically posited correspondences between relations. This allows us to identify systematic differences between the annotations. Such discrepancies could be caused by differences in the respective operationalizations in discourse annotations, or 
\textit{by implicit biases inherent in the frameworks}, and can represent valuable insight for training future automatic discourse relation classifiers.

This article first provides background on the RST-DT and PDTB frameworks (Section \ref{sec:background}), and then proceeds to laying out the proposed mappings between RST-DT and PDTB 2.0 relations according to the three recent approaches which specified such mappings \cite{chiarcos2014,bunt2016,sanders2016} in Section \ref{sec:theoretic}.  Section \ref{sec:alignment} discusses challenges due to differences in discourse segmentation between RST-DT and the PDTB, and describes the alignment algorithm for mapping RST-DT annotations to the Penn Discourse Treebank. Results of the discourse relation label mapping are discussed in Section \ref{sec:mapping}, and compared to the theoretically posited correspondences. 
Section \ref{sec:related} discusses the results from our mapping to a previous approach which used a similar methodology, albeit in a simplified setting and much smaller scale \cite{rehbein2016}.
Finally, we discuss implications for annotation as well as automatic discourse processing in Section \ref{sec:discussion}.

\noindent Our article makes the following contributions: 
\begin{itemize}
\item We propose a method for aligning RST-DT and PDTB 2.0 annotations.
\item We evaluate how well existing proposals for mapping discourse relation labels correspond to the mapping between existing RST-DT and PDTB 2.0 annotations.
\item We analyse how compatible RST-DT and PDTB 2.0 annotations are.
\item We identify sources of systematic discrepancies between annotations according to the two annotation schemes, and discuss their consequences for future annotation, corpus search, and the training of automatic discourse relation labellers.
\item We identify coherence relations for which human annotation is informative and beneficial, as well as cases for which it is unclear whether manual annotation is sufficiently consistent to be useful.
\item We provide an aligned discourse corpus where both PDTB 2.0 and RST-DT annotations can be queried simultaneously.
\end{itemize}

\section{Background}\label{sec:background}
In this section, we describe the notions underlying the two discourse relation annotation frameworks that are compared in this article, namely the PDTB 2.0 and the RST-DT, in order to provide  the necessary background for understanding the reasons behind differences in segmentation and discourse relation sense labelling that we find in our study. 

\subsection{Rhetorical Structure Theory (RST)}
\label{rst}
The framework that is used to annotate the RST-DT \cite{carlson2001} is based on the Rhetorical Structure Theory (RST) as proposed by \namecite{mann1988}. Annotators are instructed to annotate the writer's goal of each segment of a text with respect to the neighbouring segments and the resulting hierarchical structure of the entire document. 
There are different implementations of RST annotation, which follow the overall style of RST annotation, but may differ in how exactly they define the segments, called Elementary Discourse Units (EDUs) and what exact set of relation labels is chosen. 
For the current study, we focus on the RST-DT style of RST.

\paragraph{Segmentation in RST-DT}
A fundamental constraint on RST annotation is that each part of a text has to be included into the overall discourse structure, and that the discourse structure has to be arranged into a tree structure. Each document is first decomposed into non-overlapping sequential text spans (called Elementary Discourse Units). EDUs generally consist of clauses, but attributions, relative clauses, nominal postmodifiers, and phrases that begin with a strong discourse marker are also considered EDUs in RST-DT \cite[see][p.3]{carlson2001}. 

The tree structure in RST annotations does not allow crossing or embedded EDUs. In order to deal with these limitations and the restriction that EDUs cannot overlap, a \rel{Same unit} tag was introduced RST-DT, which allows annotators to express that an EDU is discontinuous. To illustrate this, consider example \ref{Example:RST-relation3} below. The clause {\em when implemented} is embedded in another clause. As a result, the clause {\em that it will} cannot be connected to its other half because {\em when implemented} cannot be skipped. The tag \rel{Same unit} can be applied in this situation to express that the segments in fact make up one unit.

\ex.
... [that it will,]  [when implemented,] [provide significant reduction in the level of debt and debt service owed by Costa Rica.] \\ --- \rel{Same-unit}, wsj\_0624
\label{Example:RST-relation3}

\paragraph{Discourse Structure in RST-DT}
RST works from the assumption that relations have at least two text elements. Example \ref{Example:RST-relation1} illustrates this: the label \rel{Elaboration-additional} was chosen to represent the semantic link between the two segments in brackets. EDUs are thus connected in a relation to form a complex discourse unit, which can itself be part of a higher-level discourse relation, until the entire text is connected and a tree structure is formed. 

Each element can either function as a {\em nucleus} or {\em satellite} in a relation. The nucleus is the central part of a text (with respect to its intentional discourse structure), while the satellite is supportive of the nucleus. 
Note that some relations have symmetrically important elements by definition. These relations consist of two nuclei rather than a nucleus and a satellite (see Example \ref{Example:RST-relation2}). 
The writer's intentions are important when assigning nuclearity (i.e., what does the writer want to achieve?). Determining nuclearity can therefore rarely be done without taking the context of the relation into consideration.
Nuclearity assignment is determined simultaneously with the assignment of a discourse relation \cite{carlson2003}.

\ex.
\a. [But even on the federal bench, specialization is creeping in,]$_{nucleus}$ [and it has become a subject of sharp controversy on the newest federal appeals court.]$_{satellite}$ \\ --- \rel{Elaboration-additional}, wsj\_0601
\label{Example:RST-relation1}
\b. [That isn't much compared with what Bill Cosby makes, or even Connie Chung for that matter (...)]$_{nucleus}$ [But the money isn't peanuts either, particularly for a news program.]$_{nucleus}$ \\ --- \rel{Contrast}, wsj\_0633
\label{Example:RST-relation2}

\namecite{carlson2003} distinguish 72 relation labels, partitioned into 16 classes that share some type of rhetorical meaning (see Appendix \ref{appA} for a list of RST-DT's relational inventory). Some of these classes contain relations that are not considered to be coherence relations in other schemes such as PDTB 2.0; examples include the {\sc Attribution} (which is also annotated in PDTB, but not considered a coherence relation) and {\sc Question-answer} relations. Another example is cases where coherence between discourse segments is not achieved through a specific semantic coherence relation but rather through cohesion. These cases of cohesion are annotated as {\sc Elaboration} relations in RST-DT.


\subsection{Penn Discourse Treebank (PDTB)-style Annotation}\label{pdtb}
PDTB-style annotation \cite{prasad2007,prasad2008,prasad2014reflections} follows a lexically-grounded approach to discourse relation representation. It distinguishes between so-called \textit{explicit} and \textit{implicit} discourse relations. Explicit relations are marked with a coordinating conjunction, subordinating conjunction or a discourse adverbial, which we will jointly refer to as \textit{discourse connectives} in this article. Implicit relations, on the other hand, are not marked with a discourse connective. Instead, annotators are asked to insert a connective they think would best fit, and annotate the coherence relation with the inserted connective. 
The framework was developed with the goal of being theory-agnostic. It therefore refrains from imposing a tree structure on annotations. 
   

\paragraph{Segmentation in the PDTB 2.0}
Relations in the PDTB have two and only two arguments, referred to as Arg1 and Arg2. These arguments can be continuous or discontinuous.
In the case of explicit relations, the argument that is syntactically bound to the connective is labeled as Arg2; the other argument is Arg1.
Annotators were instructed to first identify explicit connectives based on a list of discourse cues; they then identified the discourse relational arguments. The selection of these arguments is restricted by the ``minimality principle,'' according to which only as much material should be included in the argument as is {\em minimally required} and {\em sufficient} for the interpretation of the relation. Any other span that is considered to be relevant to the interpretation of the argument is annotated as supplementary information. After identifying explicit connectives and their arguments, a relation label is assigned, as in Example \ref{Example:pdtb-explicit}.

In PDTB 2.0, implicit discourse relations have only been annotated between adjacent sentences within paragraphs, as well as between complete clauses delimited by a semi-colon (``;'') or colon (``:'') \cite[see also][]{prasad2017}. A connective was then inserted, and the relation label assigned in a next step, see Example \ref{Example:pdtb-implicit}. 
Because arguments of implicit relations have often been annotated as complete sentences or clauses of sentences with colons or semi-colons, the annotation have a slightly different pattern in segmentation for implicit relations compared to explicit relations (this observation will become important for the design of the mapping algorithm). 

\ex.
\a. Although [that may sound like an arcane maneuver of little interest outside Washington],$_{\text{Arg2}}$ [it would set off a political earthquake].$_{\text{Arg1}}$ \\ --- \rel{Comparison.Concession.Expectation}, wsj\_0609
\label{Example:pdtb-explicit}
\b. [Mr. Carpenter denies the speculation]$_{\text{Arg1}}$ <implicit: meanwhile> [To answer the brokerage question, Kidder, in typical fashion, completed a task-force study].$_{\text{Arg2}}$ \\ --- \rel{Temporal.Synchronous}, wsj\_0604
\label{Example:pdtb-implicit}

When a discourse relation is not marked by a discourse connective, and it's also not possible to insert a suitable connective (e.g., because including the connective \conn{therefore} would sound odd and be redundant in a sentence starting with \conn{for this important reason}), the annotators can assign a coherence relation label and mark the relation as expressed via an alternative lexicalization (AltLex).

An important difference between PDTB-style annotation and RST-style annotation is that in PDTB-style annotation, it is not necessary for all parts of a text to be connected in a discourse structure. In fact, the minimality principle leads to many partial sentences not being part of any discourse relational argument. Furthermore, the restriction of only annotating implicit relations between adjacent sentences in PDTB 2.0 means that sentence-internal or cross-paragraph coherence relations may be missed (cf. \citealt{prasad2017}; but see a recent extension of PDTB annotation to VPs, \citealt{webber2016discourse}).
Additionally, PDTB differs from RST-DT in its treatment of attributions, which are text elements that ascribe beliefs and assertions to the agent(s) holding or making them \cite{riloff2003}, such as {\em he said that}. Attributions are annotated as a discourse relation in RST-DT, but they are annotated separately in PDTB \cite{prasad2007} and are not included in discourse relational arguments.

\paragraph{Discourse Structure in the PDTB 2.0}
The framework distinguishes 43 relation labels (see Appendix \ref{appA}). These labels are organised in a hierarchy consisting of three levels: (i) class is the top level, which contains the four major semantic classes; (ii) type is the second level, which further refines the semantics of the class levels; and (iii) subtype is the most fine-grained level, which defines the semantic contribution of each argument. When an annotator was uncertain of the more fine-grained senses of subtype, s/he could choose the higher level type, which was also beneficial for inter-annotator agreement \cite{prasad2008}.

A further important aspect distinguishing PDTB 2.0 annotations from RST-DT annotations is that PDTB annotators were allowed to assign several labels to the same relational arguments, when they found that multiple concurrent discourse relations held between the arguments. This is important for our evaluation of correspondences between assigned relation labels later on, as we decided to evaluate the correspondence in terms of the PDTB label that is most similar to the RST-DT label.

If no suitable connective and coherence relation was identified for two consecutive sentences, the \rel{EntRel} and \rel{NoRel} labels were used depending on whether coherence was present in the form of cohesion through shared entities or not.
The resulting annotations are not always compatible with a tree structure \cite[see][]{lee2006}. 

\section{Theory-based proposals for mapping RST-DT and PDTB 2.0 relations}
\label{sec:theoretic}

A mapping of discourse relation labels between different frameworks can be achieved by determining the correspondence of relation labels from one framework to the other directly (e.g., PDTB 2.0 to RST-DT, RST-DT to SDRT, SDRT to PDTB 2.0). Another option is to map all frameworks to an intermediary representation, such that the mapping between any two frameworks can be obtained via this intermediary representation. This approach has the advantage of being more general in case many frameworks should be mapped onto one another: rather than creating a new mapping between the new framework and all other frameworks, researchers only have to create a single mapping from the new framework to the intermediary framework. Another advantage is that it produces a candidate for a single set of relational labels potentially suitable for future annotation: the intermediary relation representation. All three of the mapping approaches discussed below propose such an intermediary representation. 

The mapping approach by  \namecite{benamara2015} could not be included here, because it focused on comparing RST and SDRT and has not yet completed the mapping to PDTB labels.\footnote{Personal communication November 2017}

\subsection{Mapping according to the OLiA reference model}
\begin{figure}[tbp]
\centering
\includegraphics[width=.6\linewidth]{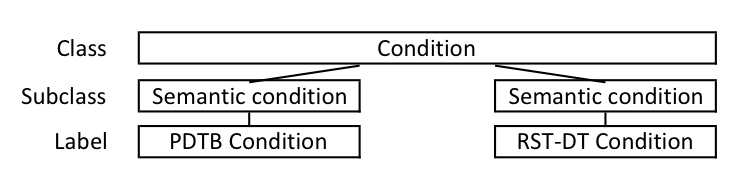}
\caption{Illustration of OLiA's hierarchical mapping approach.}
\label{fig:olia-condition}
\end{figure}

\namecite{chiarcos2014} mapped the PDTB and RST-DT schemes onto each other as part of the Ontologies of Linguistic Annotation (OLiA). OLiA provides a terminology repository that can be used to facilitate the conceptual interoperability of annotations. This is done using an intermediate level of representation that mediates between several existing frameworks. The intermediate representation is formalised as \texttt{subClassOf} descriptions. To illustrate this, consider Figure \ref{fig:olia-condition}, which illustrates a hierarchical mapping of PDTB's {\sc Condition}. In OLiA, this relation type is characterised as a subclass of {\em Semantic condition} relations, which is in turn a subclass of {\em Condition} relations. This can then be mapped onto RST-DT's class of {\em Condition}, which has the same superclasses. 
\namecite{chiarcos2014} argues that ontologies are able to represent more fine-grained nuances of meaning, and to quantify the number of shared descriptions between annotations of different frameworks \cite{chiarcos2014}.
The Table in Appendix \ref{appC} shows the proposed correspondences (indicated by a `o') between PDTB 2.0 and the RST-DT according to the proposal in OLiA.

\subsection{Mapping via Unifying Dimensions}
The Unifying Dimensions mapping (UniDim) was proposed by \namecite{sanders2016,sanderssubm} with the goal of mapping labels from different frameworks onto each other using an interlingua (see Appendix \ref{appB}). In the UniDim proposal, relation labels are not mapped to intermediate labels; rather, they are described in terms of their characteristics, or values on certain dimensions.
The set of unifying dimensions is an extended version of the dimensions originally proposed as the Cognitive approach to Coherence Relations \cite[CCR;][]{sanders1992}. 
The original CCR distinguishes four cognitive dimensions that apply to every relation, namely polarity, basic operation, source of coherence, and order of the segments. For example, a \rel{Reason} relation would be represented as a relation with positive polarity, causal basic operation, objective source of coherence, and backward order of the segments. As PDTB and RST-DT make some distinctions which cannot be represented in terms of only these four dimensions, Sanders et al.~extended CCR to account for more fine-grained properties of relations. 

The intermediate representation in terms of these dimensions allows for mapping relation labels from one framework into the representation as dimensions, and from this representation to the second framework. 
The method of describing relations in terms of their characteristics makes it easier to identify similarities and differences between relations. For example, similarities between relations can be described in terms of how many of their characteristics are identical: \rel{Cause} and \rel{Concession} differ in one dimension (polarity) but they are both types of causal, objective, non-conditional relations (in the case of \rel{Concessions}, the expected result has not occurred, or a result occurred even though the usual cause was not present). 
Appendix \ref{appC} shows the proposed correspondences (indicated by a `u') according to the proposal by \namecite{sanders2016}.

\subsection{Mapping according to the ISO standard proposal}
\namecite{bunt2016} provide a mapping of frameworks that is based on a different system. They proposed an international standard (ISO standard) for coherence relation annotation, which consists of a set of 20 core relations that are commonly found in some form in existing approaches (see Appendix \ref{appB}). They did not aim to provide a fixed and exhaustive set of coherence relations; rather, they aimed at providing an open, extensible set of relations.
\namecite{bunt2016} propose that the ISO standard can be used for future annotation efforts, as well as for mapping between annotations using different frameworks.
To this end, they provided mappings of these ISO relations to other frameworks, including PDTB and RST-DT.\footnotetext{Mappings from the paper for certain labels were updated; personal communication January 2018.} From these proposed correspondences to ISO standard candidate relations, we can infer how the PDTB 2.0 and RST-DT relations correspond to one another. The full table of hypothesized correspondences according to the ISO proposal are marked by `i' in Appendix \ref{appC}.



\subsection{Discussion of agreement and discrepancies between proposed mappings}
Appendix \ref{appC} shows the grid of proposed correspondences between the schemes.  While we can see many cells that include `o' , `u' and 'i', indicating that all approaches agree that these relation labels should correspond to one another, we can also see some relational labels that have only been linked by one of the proposals. 
We have identified three main reasons for these discrepancies, which we discuss below: differences in granularity of mapping schemes, differences in how concepts (in particular, order and subjectivity) are defined by the frameworks, and differences in the interpretation of definitions in the annotation guidelines. 

\paragraph{Granularity of the proposals}
The first source of discrepancies is the granularity of the intermediate schemes. For cases where ISO is coarser than the relational categories included in RST-DT and PDTB 2.0, the mapping differs from OLiA's and UniDim's mapping.
For example, the causal relation labels in the ISO-based mapping are coarser than the other two schemes, because the proposed ISO standard doesn't differentiate between semantic and pragmatic causal relations (a distinction also known as objective vs.~subjective or content vs.~epistemic); i.e.~causal relations where the link can be established based on the semantics of the two arguments and causal relations where the author posits a causal link. OLiA and UniDim do distinguish between semantic and pragmatic relations, and therefore do not map certain RST-DT labels that the manual mentions are semantic (e.g., {\sc Reason} and {\sc Explanation-argumentative}) to PDTB's pragmatic causal label {\sc Justification}, whereas ISO does map semantic RST-DT labels to {\sc Justification}.

An example of a relational category that is more fine-grained in the ISO proposal than in the frameworks is {\sc Exception}, which corresponds to PDTB's {\sc Exception}, but has no equivalent mapping to an RST-DT relation. As a result, PDTB's {\sc Exception} cannot be mapped to a corresponding RST-DT label based on the ISO proposal. In UniDim, however, \rel{Exception} is mapped to \rel{Contrast}, \rel{Antithesis} and \rel{Preference}, i.e.~a set of more general labels with similar characteristics to PDTB's \rel{Exception}. The same goes for RST's {\sc Means} relations.

Certain labels can also not be mapped by OLiA. For example, RST-DT's {\sc Evaluation}, {\sc Comment} and {\sc Definition} relations are part of the superclass {\em Assessment}, which doesn't occur in the PDTB inventory; furthermore, the relations {\sc Background} and {\sc Circumstance} are part of the superclass {\em Background}, which also doesn't occur in PDTB. OLiA also doesn't map any  {\sc Comparison} RST-DT relations. Labels could be mapped using their supersuperclass, but this would not be very informative (e.g., it would result in mapping the {\em Background} label to all PDTB {\sc Expansion} labels).

In temporal relations marked with connectives such as \textit{before} and \textit{after}, PDTB assigns the label \rel{Temp.Async.Succession} to relations marked with \textit{after}, and the label \rel{Temp.Async.Precedence} to relations marked with \textit{before} as a subordinating conjunction. Similarly, RST uses the labels \rel{Temporal-after} and \rel{Temporal-before} for these instances. UniDim and OLiA therefore proposed to map \rel{Temp.Async.Succession} onto \rel{Temporal-after} and \rel{Temp.Async.Precedence} to \rel{Temporal-before}. 
ISO on the other hand more generally maps both of PDTB's \rel{Temporal.Async} relations onto all asynchronous temporal relations in RST. 

Discrepancies in granularity occur because of differences in the goals of the mapping frameworks, and different systems of mapping. UniDim was designed to be able to map labels, and, as a result, the interlingua can be used to map relational labels even when there is no direct equivalent. 
ISO, on the other hand, was proposed as a new set of relations; i.e.~mapping was not its primary goal. The mapping that is provided in \namecite{bunt2016} mainly includes direct correspondences. OLiA also has the goal to identify all correspondences between frameworks, but because of the hierarchical structure of the interlingua, a relation label that is part of a unique superclass in one framework can often not be mapped properly to another framework.

\paragraph{Subjective vs.~epistemic relations}



A somewhat difficult issue is the notion of subjectivity. A relation can be objective, subjective or epistemic. RST distinguishes between objective and subjective relations with labels such as \rel{Cause} and \rel{Result} on the one hand and \rel{Reason}, \rel{Evidence} and \rel{Explanation-argumentative} on the other hand, while PDTB distinguishes between non-epistemic relations \rel{Contingency.Cause.Reason} and \rel{Contingency.Cause.Result} and epistemic relations such as \rel{Contengency.Pragmatic~Cause.Justification}. 
OLiA proposed to map RST-DT's subjective labels such as {\sc Evidence} label only to PDTB's epistemic label {\sc Pragmatic cause}, which does however not do justice to PDTB's more restrictive notion of subjectivity as epistemic relations only. A similar problem occurs for RST-DT's {\sc Manner, Means, Problem-Solution} and \rel{Enablement}.

\paragraph{Difference in interpretations of definitions}
The third source of discrepancies is rooted in different interpretations of relational definitions in the annotation manuals, and hence represents the theoretically most interesting case for comparison between proposals. In the remainder of this section, we will focus our discussion on these cases\footnotetext{Differences that are not discussed in this section include for instance: UniDim's mapping of RST-DT {\sc Hypothetical} to PDTB {\sc Pragmatic condition} (in addition to {\sc Condition}); ISO's mapping of RST-DT \rel{Evaluation} to the general \rel{Expansion} class; OLiA's mapping of RST-DT \rel{Summary} relations to \rel{Equivalence} (in addition to \rel{Specification} and \rel{Generalization}).}.
%
%
Table \ref{tab:discrepancies} provides an overview of the differences between the OLiA, UniDim and ISO-based mapping according to this type of discrepancies.

\begin{table}
\centering
\caption{Overview of the differences between the OLiA, UniDim and ISO-based mappings caused by different interpretations of definitions.}
\setlength{\tabcolsep}{5pt}
\begin{tabular}{l|lll}
\hline
\multirow{2}{*}{{\bf RST-DT label}}  & \multicolumn{3}{c}{{\bf Mapping to PDTB according to:}  }     \\
                      & {\bf OLiA}    & {\bf UniDim   }                  & {\bf ISO   }                    \\ \hline
\rel{Comparison}  &  Not mapped   & \rel{Conjunction}        & \rel{Contrast}          \\ \hline
\multirow{2}{*}{\rel{Antithesis}  }  &  \rel{Contrast} & \rel{Concession}         & \rel{Concession}        \\ 
                     &    & \& \rel{Contrast}        &                           \\ \hline
\rel{Elab.-object-} &  \rel{Expansion}  & \rel{Specification}      & \rel{EntRel}            \\
\rel{attribute} & & \& \rel{Generalization}  &         \\ \hline
\multirow{2}{*}{\rel{Background}  } &  Not mapped  & \rel{Conjunction}      & \rel{Conjunction}            \\ 
 &     & \rel{\& Asynchronous}      &       \\  \hline
\multirow{2}{*}{\rel{Circumstance}  } &  Not mapped  & \rel{Conjunction,}      & \rel{Synchronous}            \\ 
 &     & \rel{Synchronous}      &       \\  
  &     & \rel{\& Asynchronous}      &       \\  \hline
\end{tabular}
\label{tab:discrepancies}
\end{table}

\paragraph{RST-DT's {\sc Comparison}}
First, the proposals differ in their mapping of RST-DT's {\sc Comparison} relations. The manual states that the two segments of a {\sc Comparison} relation are not in contrast with each other \cite[][p. 50]{carlson2001}. Based on this description, UniDim mapped \rel{Comparison} to PDTB {\sc Conjunction}. ISO, however, mapped \rel{Comparison} to PDTB's {\sc Contrast} relational class. Finally, OLiA mapped RST-DT's \rel{Comparison} to the superclass {\em Non-contrastive comparison}. However, none of the labels in the PDTB have a superclass {\em Non-contrastive comparison}, and therefore RST-DT's labels do not have a correspondence in PDTB.
The mapped data will be able to indicate how the \rel{Comparison} label was used in practice

\paragraph{RST-DT's {\sc Antithesis}}
The frameworks also disagree on the mapping to contrastive and concessive relations. The former is a relation of semantic opposition; the latter contains a denial of expectation. 
To illustrate the difference between the two types of relations, consider the following examples:

\ex.
Dylan used to live in Washington DC, but now he lives in Baltimore.
\label{ex:contrast}

\ex.
Dylan lives in Baltimore, but he works in Washington DC.
\label{ex:concession}

Example \ref{ex:contrast} presents a simple contrast between where Dylan used to live and where he lives now. The relation could also have been expressed by the connective \conn{whereas}, which is a typical marker of contrastive relations. In Example \ref{ex:concession}, the first segment informs you where Dylan lives now. This segment presupposes an implicit expectation that Dylan also works there, but the second segment denies this expectation: Dylan works in a different city. This is typical of \rel{Concession} relations: one argument creates an expectation of a cause or consequence, which is denied by the other argument. The relation could also have been expressed with  typical markers of concessive relations, such as \conn{even though} or \conn{nevertheless}. The distinction between \rel{Contrast} and \rel{Concession} is relatively difficult to make, even for trained annotators \cite[see, for example,][]{robaldo2014,zufferey2013}.

The frameworks all disagree on the mapping of the RST-DT label \rel{Antithesis} to PDTB {\sc Contrast} and {\sc Concession}. 
RST-DT's annotation manual states that {\sc Antithesis} is a contrastive relation, but some of the examples that are provided are concessive relations. In UniDim, {\sc Antithesis} is therefore mapped to both contrastive and concessive PDTB relational labels, whereas in ISO, {\sc Antithesis} is mapped only to concessive labels. In OLiA, \rel{Antithesis} is mapped onto \rel{Contrast} only.


\paragraph{RST-DT's {\sc Elaboration-object-attribute}}
The proposals differ in their mapping of RST-DT's {\sc Elaboration-object-attribute} label: OLiA maps it to general {\sc Expansion} relations, UniDim maps it to PDTB's {\sc Specification} and {\sc Generalization} relations, while the ISO-based proposal maps {\sc Elaboration-object-attribute} to {\sc EntRel}. 

\paragraph{RST-DT's {\sc Background} and {\sc Circumstance}}
Finally, UniDim and ISO differ in their treatment of RST-DT's {\sc Background} and {\sc Circumstance}. The two proposals agree on the mapping of \rel{Background} to PDTB \rel{Conjunction} relations, but UniDim also maps \rel{Background} to \rel{Temporal.Asynchronous} based on the description of \rel{Background} in the manual: ``the events represented in the nucleus and the satellite occur at distinctly different times'' \cite[p. 47]{carlson2001}. 
Regarding \rel{Circumstance} relations, UniDim and ISO agree on mapping these to \rel{Synchronous} relations, but UniDim also maps to \rel{Asynchronous} and \rel{Conjunction} labels.

\bigskip


The disagreements that are attributable to different interpretations of the definitions in the manual will be systematically analysed with respect to actual annotations, and discussed in  Section \ref{sec:mapping}, in order to determine which of the proposed correspondences are justified by the actual data. These results can then also be used to clarify annotation guidelines for future research, and highlight which definitions are particularly susceptible to inconsistent annotation.

\section{Data, Segmentation and Automatic alignment}\label{sec:alignment}
PDTB 2.0 and RST-DT annotations overlap for 385 newspaper articles in sections  6, 11, 13, 19 and 23 of the Wall Street Journal corpus. 
The annotation of the RST-DT involved more than a dozen of people and several phases of revision. The average inter-annotator agreement (final results for 6 taggers) on span detection, nuclearity assignment and relation sense annotation was 86.8\%, 80.7\%, and 72\%, respectively \cite{carlson2003}.\footnote{These numbers are the result of averaging over the inter-annotator agreement scores reported for every two annotators in Table 2 of \namecite{carlson2003}.}
The PDTB 2.0 reports an inter-annotator agreement of 94\%, 84\%, and 80\% for the class, type and subtype levels respectively, and
PDTB's discourse segments were identified with an agreement (exact string match) of 90.2\% for explicit relations and 85.1\% for implicit relations \cite{prasad2008}.

Our investigation will be based on the intersection of the PDTB and the RST-DT, with annotations from both frameworks included as different annotation layers.

\subsection{Segmentation}
Comparing annotations of the two corpora is not a trivial task, because annotations not only differ in the label sets that were used, but also in terms of segmentation.
Firstly, there are discrepancies in what is considered an "elementary discourse unit" in RST-DT vs.~what is considered a discourse relational argument or an attribution in PDTB 2.0. Note that in the remained of this paper, we use the term ``segment'' to refer to the text elements that are part of a relations; that is, the {\em arguments} in PDTB and the {\em EDUS}, {\em nuclei} and {\em satellites} in RST-DT.
There are also differences in the discourse structure: RST annotates discourse trees spanning the whole document, while PDTB 2.0 only annotates relations between adjacent sentences and relations marked by an explicit connective. 
This results in a considerably lower number of PDTB relations than RST-DT relations for the same text. We therefore use PDTB relations as a starting point in alignment, with the goal of identifying for each PDTB relation the corresponding relation label in the RST annotation.

PDTB's minimality principle (cf.~Section \ref{pdtb}) and RST's tree structure (cf.~Section \ref{rst}) influence the result of the segmentation and annotation steps.
An annotation alignment process must therefore take into account systematic differences arising from the respective segmentation procedures. In the automatic alignment step, our goal is to map as many discourse relation labels as possible in order to get a maximally complete picture regarding how well the annotations correspond to one another. At the same time, we must only map those labels where annotators inferred the same relation -- if the RST-DT annotators annotated a relation holding between two text segments, and the PDTB annotators marked a relation between two different segments, these labels should \textit{not} be recorded as valid alignments, and labels hence shouldn't be compared. To illustrate this, consider Example \ref{fig:segmentation}, which presents the PDTB (left) and RST-DT (right) annotations for a fragment of a Wall Street Journal article. PDTB segmented Arg1 differently than RST-DT, leading to a difference in interpretation. PDTB considers segments (c-d) as the result of the event in segment (b), whereas RST-DT focused on the different opinions expressed in segments (a-b) and (c-d). The disagreement between the two labels (\rel{Result} vs.~\rel{List}, respectively) does not stem from annotator disagreement regarding the label, but from a more fundamental difference in segmentation. Such cases should therefore not be included in the evaluation of mapped labels.

\begin{figure}  
\centering
\includegraphics[width=0.9\linewidth]{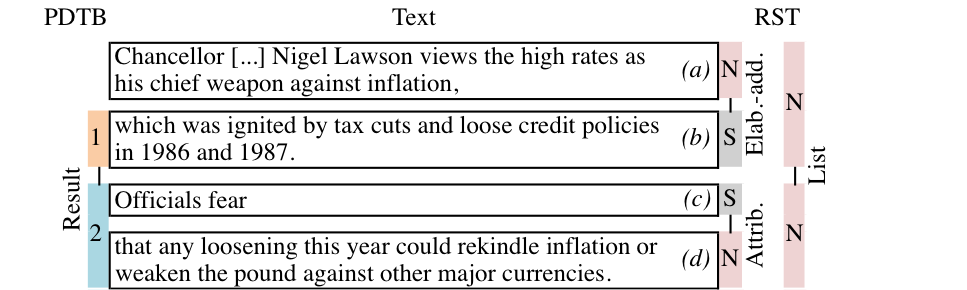}
\caption{PDTB and RST-DT annotations for a paragraph of wsj\_1172. {\em 1} refers to Arg1 in PDTB; {\em 2} refers to Arg2. {\em N} refers to Nucleus in RST; {\em S} refers to Satellite. (a-d) refer to RST-DT's EDUs.}
\label{fig:segmentation}
\end{figure}

\begin{figure}[tbp]
\includegraphics[width=\linewidth, trim= 0cm 1.34cm 0cm 0cm, clip=TRUE]{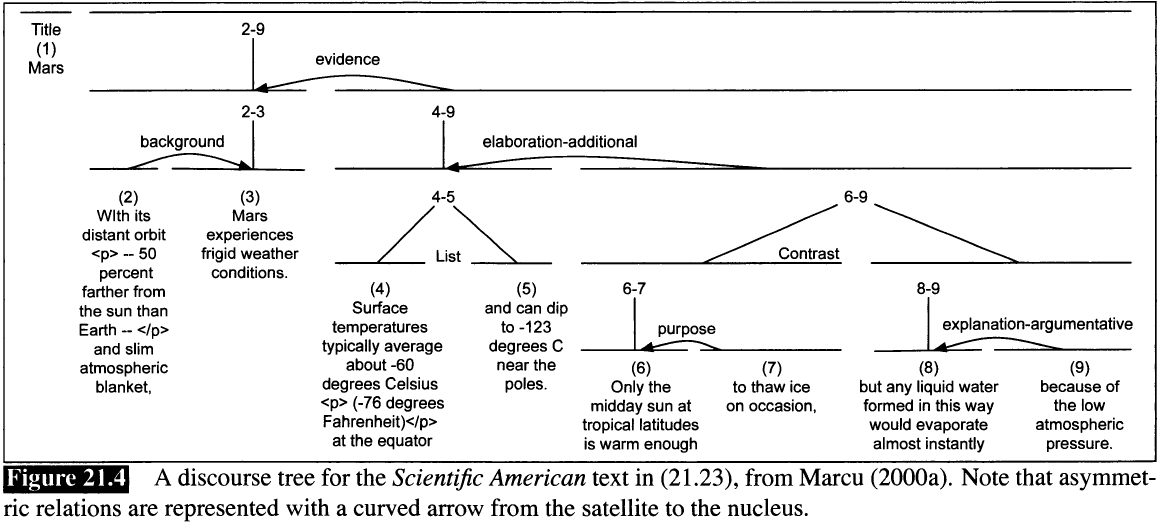}
\caption{RST discourse structure, Figure 1.1 from \namecite{marcu2000}.}
\label{fig:seg1}
\end{figure}

In related work, \namecite{scheffler2016konvens} compared PDTB 3.0-style annotations and RST-style annotations on the German Potsdam Commentary Corpus \cite[PCC;][]{stede2014potsdam}. 
The PCC includes 1104 explicit connectives annotated in PDTB-style but lacks PDTB-style annotations for implicit relations. \namecite{scheffler2016konvens} propose a simple method for mapping the PDTB 3.0-style and RST-style annotations for explicitly marked relations in the Potsdam Commentary Corpus (PCC) onto one another, in order to empirically observe commonalities and differences in the annotations of discourse structure between the two approaches. They do not, however, compare relation labels for the mapped relations. The alignment algorithm used in \namecite{scheffler2016konvens} compares the spans of the relation argument annotations, and distinguishes different segmentation constellations. They observe that the majority (84\%) of instances in their corpus consists of cases that are easy to map, including exact match of discourse relational arguments (41\%) and ``boundary match'' (39\%), where a relation is annotated between two adjacent text spans, and the boundary between the two arguments is identical. They however also report difficult cases of non-local relations where the segment boundaries differ (13\%) (as in the example of the contrast relation in Figure \ref{fig:seg1}), and cases in which no match is possible (3\%). 

The alignment algorithm proposed in this article (see Section \ref{sec:alg} below) is more complex than the one proposed in \namecite{scheffler2016konvens}, in order to better address those cases for which there are differences in segmentation between annotation layers. The core idea of how valid alignments can be identified even in the face of mismatches between relational arguments builds on the Strong Nuclearity hypothesis \cite{marcu2000}, which was used for RST-DT annotation. The Strong Nuclearity hypothesis states that when a relation is postulated to hold between two spans of text, it should also hold between the nuclei of these two spans. Note that the notion of ``nuclearity'' has had several slightly different interpretations throughout the conception and further development of RST, as laid out in \namecite{stede2008rst}. Independent of the theoretical discussion about the intentions behind nuclearity as such, the specific notion used in RST-DT annotation is helpful for determining relation alignment. RST-DT's nuclearity was assigned in an instance-by-instance decision for identifying those parts of a discourse relation which are crucial for that relation to hold; it was not used as a general property of relation types, as in some other variants of RST annotation. In that sense, the strong nuclearity annotation guideline from \namecite{marcu2000} is related to the minimality principle used in PDTB 2.0 annotations: both help to indicate which segments of the text are central to establishing the discourse relation.  
To illustrate this, consider the {\sc Contrast} relation in Figure \ref{fig:seg1}; the Strong Nuclearity hypothesis means that if the relation holds between (6-7) and (8-9), it should also hold between (6) and (8), but not between (7) and (8) or (7) and (9). In the following, we will use the expression \textit{nucleus path} to refer to the path between a complex argument of a high-level relation, and the single EDU which one ends up with if always following the path down the segments annotated as nucleus.

\subsubsection{Segmentation constellations}
We will now go through the different segmentation and alignment constellations using examples, to explain where challenges in alignment lie, and how these are dealt with by our alignment procedure. For ease of reference, we will here adopt the PDTB distinction between explicitly marked and implicit relations, even when referring to RST-DT annotations.

\paragraph{PDTB relations with adjacent arguments} 
The simplest case is an exact match between the discourse relational arguments for the two annotation layers.
Additionally, there can be cases where the argument spans largely overlap but differ in their exact boundaries.
Consider Figure \ref{fig:segmentation}: in PDTB 2.0, segments (a-b) and (c) are connected in a {\sc Temporal.Synchrony} relation.
In RST-DT, a {\sc Temporal-same-time} relation was annotated between segments (b) and (c). Even though the spans differ in whether (a) is included, they clearly correspond to each other. As neither of the discourse relational arguments is complex in the case of such matches, it is straightforward to decide which relation labels should correspond to each other. 



\begin{figure}  
\centering
\includegraphics[width=0.9\linewidth]{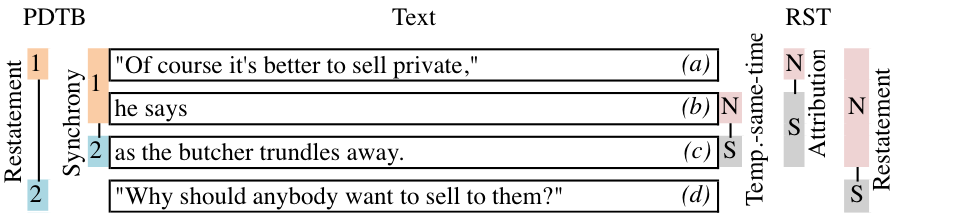}
\caption{PDTB and RST annotations for a section of wsj\_1146.}
\label{fig:segmentation}
\end{figure}

%

\paragraph{PDTB relations with non-adjacent arguments}
We also frequently encounter more complex cases, where the PDTB arguments are not directly adjacent to one another. This can happen both for explicitly marked relations and for implicit relations (when the sentences are adjacent but the chosen spans do not cover the complete sentence). Whenever the PDTB arguments are not adjacent, we will either have a mismatch between the size of the discourse segments in that the RST-DT EDU is larger than the PDTB argument, or in that the RST-DT argument is \textit{complex}, i.e.~it consists of other relations.
In Figure \ref{fig:segmentation}, this is the case for the {\sc Restatement} relation: in PDTB, this relation holds between segments (a) and (d), whereas in RST-DT, the relation holds between segments (a-c) and (d). For deciding whether the PDTB \rel{Expansion.Restatement} and RST-DT \rel{Restatement} relations should be aligned, we rely on the Strong Nuclearity hypothesis. It says that the complex relation between (a-c) and (d) should also hold between the nucleus of (a-c), hence (a) and (d). We can then infer an exact match between discourse relational arguments (a) and (d) between the two annotation layers, and map the labels onto one another. 

Such cases also occur among explicitly marked relations. Since RST-DT relations are annotated in a hierarchical tree structure, relations connecting non-adjacent sentences will have large discourse relational arguments. Consider Example \ref{fig:seg1} again: RST-DT annotates a {\sc Contrast} relation with segments (6-7) as one nucleus and (8-9) as the other nucleus. In the PDTB annotation, because of the minimality principle of marking discourse relational arguments, the relation marked by \textit{but} has segment (8) as its ARG2 and segment (6) as its ARG1. Similarly, the {\sc evidence} relation between segments (2-3) and (4-9) would differ in terms of its argument boundaries in PDTB annotation, as segments (6-9) would typically not be included in the ARG2 of the implicit relation. Nevertheless, an alignment of relations is possible given nuclearity annotation, and labels from such cases are included in the mapping.

\paragraph{Relations with inconsistent nuclearity}
There are however also cases for which labels should not be mapped due to discrepancies in what relation the annotators intended to label. These cases can typically be identified through inconsistencies between the discourse relational arguments annotated by PDTB and the nuclearity assignment annotated in RST-DT. 

\begin{figure}  
\centering
\includegraphics[width=0.9\linewidth]{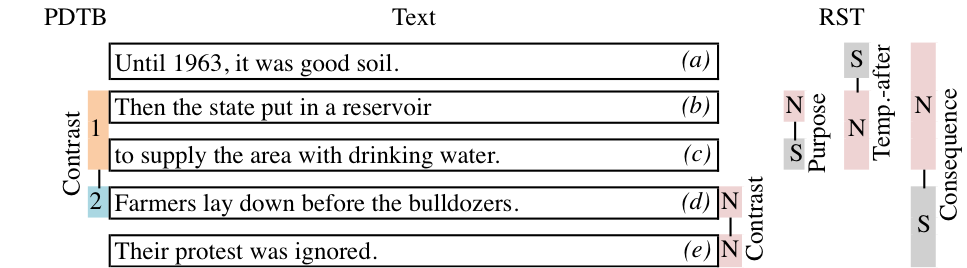}
\caption{PDTB and RST-DT annotations for a paragraph of wsj\_1146. Note: Only the PDTB annotation that is relevant for this example is included.}
\label{fig:revise-bad}
\end{figure}

To illustrate this, consider the passage in Figure \ref{fig:revise-bad}. 
In this example, one would have to map PDTB relation \rel{Contrast} to RST-DT's \rel{Consequence}, if one were to only take into account maximal overlap of discourse segments, but ignore nuclearity: the difference in span size (PDTB's annotation excludes segments (a) and (e)) affects the interpretation of the relation. In PDTB's annotation, the relation holds between the state's action and the farmers' actions. RST-DT's annotation, on the other hand, the annotated relation holds between the state's action and the consequences of that action. The relation labels therefore correspond to different interpretations. 
Such cases can be identified automatically because PDTB's Arg2 cannot be safely traced to the nucleus of the satellite of RST's \rel{Consequence} relation, as the contrast relation annotated in RST is multinuclear. Such cases are automatically flagged by our algorithm and excluded from the mapping analysis.

In order to investigate how often relations with intervening multinuclear relations (i.e., relations in which one of the segments consists of a larger tree branch including a multinuclear relation) occur in the data and to what extent they pose a problem for automatic alignment, we extracted all instances that contain an intervening multinuclear relation. In total, 892 relations (13\% of the data) have one or more intervening multinuclear relations. However, not all of these pose a potential risk to alignment -- cases where the multi-nuclear relation is the relation to be compared, and multi-nuclear relations that are not on the nucleus path, are safe to map.
We found that 295 multinuclear cases were flagged as potentially violating strong nuclearity. We manually checked 50 of these flagged instances and found that almost all of them indeed do not represent valid alignments, and should therefore be excluded from the mapping analysis.


\paragraph{Internal relations}
The segmentation granularity between the two frameworks can differ, which can lead to a specific instance from the finer-grained framework not being mapped to a label in the coarser framework. For example, two RST-DT EDUs could occur internally within a sentence, without this relation being annotated in the PDTB annotation layer.
Figure \ref{fig:seg1} shows an example where the {\sc Background} relation between segments (2) and (3) has no corresponding PDTB annotation. Similarly, there are also cases where a discourse adverbial is annotated as an explicit marker for a relation in the PDTB 2.0 annotation, but both arguments of this relation are part of the same RST EDU, as in Example \ref{ex:pdtbfinegrained}: PDTB annotated the connective {\em because} whereas in RST-DT, the entire sentence was considered as one EDU. 
Internal relations can inherently not be aligned, and hence were excluded from our mapping analysis.

\ex. [That's] because [municipal-bond interest is exempt from federal income tax -- and from state and local taxes too, for in-state investors.]  \\ --- \rel{Contingency.Cause.Reason}, wsj\_0689
\label{ex:pdtbfinegrained}

\begin{figure}  
\centering
\includegraphics[width=0.9\linewidth]{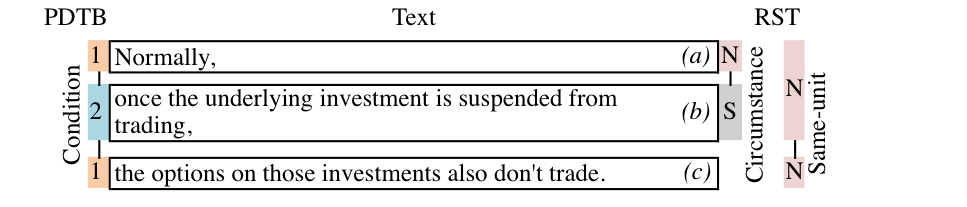}
\caption{PDTB and RST-DT annotations for a paragraph of wsj\_1962.}
\label{fig:sameunit}
\end{figure}

\paragraph{RST-DT's \rel{Same-unit}}
Centrally embedded relations (where the ARG1 is discontinuous and ARG2 is located inside the span of ARG1) can be successfully mapped by our algorithm if the RST-DT annotation contains a {\sc Same-Unit} annotation that links the discontinuous parts of the RST segments mapping to the PDTB ARG1.
This is illustrated in Figure \ref{fig:sameunit}. If a {\sc Same-Unit} annotation in RST-DT is identified, the label of the corresponding PDTB relation is mapped to the label of the relation below the \rel{Same-unit} relation (96 instances). We manually verified a subset of mapped relations to make sure that this heuristic is valid in practice (see Section \ref{sec:align-results}). For \rel{Same-unit} relations where both of the RST-DT segments contain multiple EDUs (44 instances), the corresponding relation could not be determined. These cases are flagged and excluded from the mapping analysis.


\paragraph{Cohesion}
The PDTB relation label \rel{EntRel} is used to mark cohesion when no specific coherence relation can be identified. As cohesion is not defined as a type of coherence relation, the \rel{EntRel} label is not included in the UniDim mapping. However, RST-DT contains several labels (\rel{Elaboration-additional}, \rel{Definition}, \rel{Background}) that tend to be annotated when there is cohesion but no easily identifyable coherence relation. In our analysis, we therefore decided to include the \rel{EntRel} label, to check whether it actually coincides with RST-DT's labels signalling cohesion.


\subsection{Alignment Algorithm}
\label{sec:alg}
Our procedure for mapping annotations provided by the two frameworks takes the PDTB relations as a starting point, because there are fewer relations annotated in the PDTB 2.0 than in the RST-DT. The alignment procedure is aimed at determining the optimal mapping of each PDTB relation to an RST relation. Thereby, our goal is to identify as many valid correspondences between annotations as possible, but at the same time minimise mapping ``noise'' by also identifying those cases in which we cannot be sure that the annotators identified the same underlying relation.\footnote{The alignments will be made available.}
Our mapping algorithm involves two major steps:
 
 \begin{enumerate}
\item Identifying for every PDTB discourse relation those RST-DT segments (EDUs or sub-trees containing more than one EDU) that best correspond to the PDTB segments Arg1 and Arg2 separately. \\
\item	Identifying the RST-DT relation label that describes the relation between the Arg1-equivalent and Arg2-equivalent spans. \\
\end{enumerate}

In the first step, for each PDTB argument, we iterate over all RST EDUs in the source file and select the one with maximum overlap (common characters) and minimum margin (extra characters). 
We then determine whether a PDTB argument should be aligned to more than a single RST EDU by iterating over all RST relation annotations (sub-trees spanning over several EDUs) using the same criteria. Having identified the closest matching RST-annotated text spans for both arguments of the PDTB relation (Arg1-equivalent and Arg2-equivalent RST spans), we move to step 2 to find the lowest RST relation within the discourse tree that contains the two RST spans obtained in the previous step in different arguments. In this step, several error flags are set for manual investigation of possibly invalid alignments, which we briefly introduced in the previous section and will discuss in more detail below.

\begin{figure}  
\centering
\includegraphics[width=0.9\linewidth]{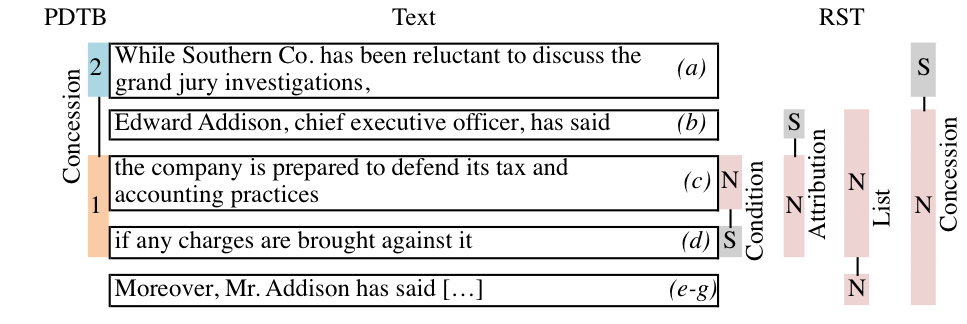}
\caption{PDTB and RST-DT annotations for a paragraph of wsj\_0619. Note: Only the PDTB annotation that is relevant for this example is included. Segment (e-g) is simplified, it actually consists of multiple RST-DT relations.}
\label{fig:difficultsuccess2}
\end{figure}

To illustrate the alignment procedure, consider the example shown in Figure \ref{fig:difficultsuccess2}. The first step of the alignment algorithm is to identify the corresponding text segment for PDTB's Arg2 (here, segment a) and Arg1 (here, segments c-d). In order to do that, the algorithm would first compare all EDUs separately to the PDTB argument spans to identify the ones with most character overlap, and then move on to comparing larger spans. This would mean that it would first identify (c) as a matching EDU for PDTB's Arg1, and then replace this by the better-matching combination of the span comprising both EDUs (c) and (d). 

We can see that in this example, segments (c-d) are conjoined in a \rel{Condition} relation in RST-DT, and are part of an \rel{Attribution} relation with segment (b), as well as a \rel{List} relation with segments (e-g). Finally, there is a relation connecting segment (a) to segments (b-g). PDTB's Arg1 is hence embedded in several relations as part of a tree branch in RST-DT, including one multinuclear relation (the \rel{List} relation). 

As part of its second step, the automatic algorithm proposes an alignment of the PDTB \rel{Concession} label to the RST \rel{Concession} label, because RST's \rel{Concession} relation is the lowest relation which includes the PDTB Arg2-equivalent span (a) and the Arg1-equivalent span (c-d) in separate arguments. However, it would also automatically flag this instance due to the intervening multi-nuclear \rel{List} relation: the multi-nuclear relation prevents it from unambiguously identifying the nucleus path from the nucleus of RST's \rel{Concession} relation to the relevant textspan that corresponds to PDTB's Arg1.


%

The algorithm flags all instances for which the mapping was potentially problematic, based on various criteria. 
First, relations for which PDTB arguments are discontinuous (e.g., containing text spans which are marked as not belonging to the relation, or centrally embedding the other PDTB argument, or overlapping with it) are flagged to allow for further manual checking.
Second, relation mappings that are inconsistent with the Strong Nuclearity hypothesis \cite{marcu2000} were flagged so that these cases could be excluded from mapping. 
Additionally, we added flags for relations that contain intervening multi-nuclear relations, for relations that were originally labelled as {\sc Same-unit}, and for relations that contain an intervening RST-DT {\sc Attribution} relation.


\subsection{Results of the alignment procedure}\label{sec:align-results}
In total, we were able to include 76\% of PDTB relations from the joint corpus into our mapping analysis (a total of 5141 relations). 52\% of these relations (a total of 2662) have directly corresponding argument spans, for which argument spans are exactly identical or differ only with respect to punctuation or inclusion/exclusion of connective in a segment. 

The remaining 48\% (2489 instances) of the data included in the mapping analysis consists of relations for which the RST-DT tree is more complex than the PDTB relation. In other words, at least one of the PDTB arguments mapped onto an RST-DT relation that consists of multiple RST-DT EDUs.
In order to investigate whether these more complex relations are aligned correctly, we randomly selected 100 instances and evaluated whether the algorithm was justified in mapping PDTB and RST-DT labels for these instances. We found that 95 relations were mapped successfully, while 5 instances were unjustified. In these five cases, the nucleus of a larger RST-DT span matched PDTB's argument, but the annotators did not evaluate the same type of relation. This was largely due to one of the segments (usually Arg2) being part of a larger branch in RST-DT. Even though this segment was the nucleus of that branch, it still blocked a stronger interpretation.

\begin{figure}  
\centering
\includegraphics[width=0.9\linewidth]{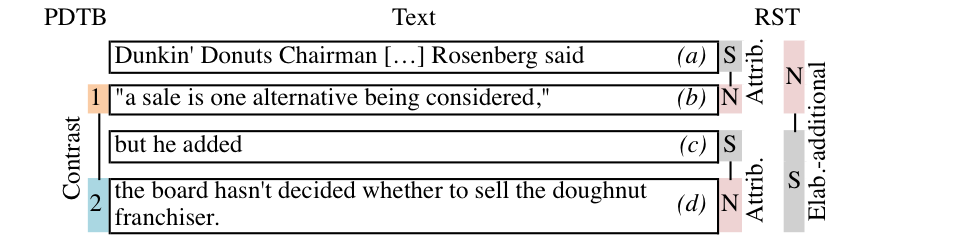}
\caption{PDTB and RST-DT annotations for a paragraph of wsj\_1176. Note: Only the PDTB annotation that is relevant for this example is included.}
\label{fig:unjustified-okay}
\end{figure}

To illustrate this, consider the example in Figure \ref{fig:unjustified-okay}.
PDTB annotated a \rel{Contrast} relation between segments (b) and (d), whereas RST-DT annotated an \rel{Elaboration-additional} relation between segments (a-b) and (c-d). The inclusion of the two attributions changes the interpretation of the two segments: the relation is focused on the speaker saying something and then adding to that (expressed by an \rel{Elab.-additional} relation), rather than on the content of what the speaker is saying (expressed by a \rel{Contrast} relation). Although both the PDTB and the RST-DT annotators selected the same arguments for these relations, the labels might not always match because {\sc Attribution} can (in a subset of cases) ``block'' an interpretation. To quantify the risk related to intervening attributions, we counted the occurrence of attribution relations in the otherwise successfully mapped relations and found that 595 relations (12\%) in total have at least one attribution relation in one of the segments; 49 (less than 1\% of the data)  have two or more intervening attributions (like the example given above). Note that out of these 49 instances, only some exhibit the problem of attribution leading to different annotated labels; we therefore decided to include these instances in our analysis. 

We conclude that our manual inspection of a representative sample of instances confirms that our alignment algorithm is highly reliable.


\subsubsection{Instances that could not be successfully aligned}
24\% (1621 instances) of PDTB relations were flagged by at least one of our flags indicating difficult cases. That is, these relations were automatically aligned, but these instances exhibit e.g.~inconsistent nuclearity, such that there is a higher chance that the annotated relation labels do not correspond to one another (i.e., annotators had different discourse structures or interpretations in mind). 

To provide a more quantitative idea of the cases that were excluded from the mapping, we again randomly selected and manually evaluated 100 instances. 
We found that 72 items were correctly flagged as unjustified mappings. Out of these, 
15 instances consisted of \rel{EntRel} labels. Finding many \rel{EntRel} labels among the flagged instances is expected, since \rel{EntRel} is annotated between adjacent segments that do not have a stronger reading. Often, these occur on the boundary of larger RST-DT spans. Looking at the total set of flagged mappings, we find that \rel{EntRel} makes up a large portion of this set: it occurs 333 times, which amounts to 19\% of all flagged mappings. Additionally, PDTB's \rel{NoRel} occurs 32 times in the flagged mappings ({\sc NoRel} is used for adjacent arguments between which no discourse relation holds). 
The 72 correctly flagged relations also included four cases of \rel{Same-unit} relations which could not be mapped automatically because both segments of these Same-unit relations consisted of multiple EDUs. 

An additional four \rel{Same-unit} relations were correctly mapped by choosing the relation label below the \rel{Same-unit} relation. After inspection of additional samples of \rel{Same-unit} relations for which the label could be unambiguously resolved automatically, we decided to include all of the unambiguously resolvable \rel{Same-unit} relations into our analyses, while \rel{Same-unit} relations where both segments have multiple EDUs are excluded.

\begin{figure}  
\centering
\includegraphics[width=0.9\linewidth]{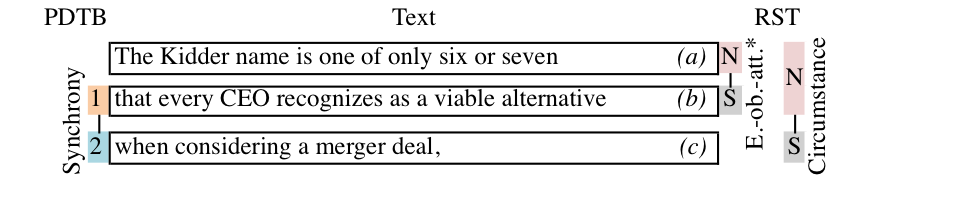}
\caption{PDTB and RST-DT annotations for a paragraph of wsj\_0604. Note: Only the PDTB annotation that is relevant for this example is included. *The full label is {\sc Elaboration-object-attribute}.}
\label{fig:badalignment}
\end{figure}

The remainder of 24 relations do in fact represent valid mappings. This illustrates that our algorithm prefers high reliability of the mapping over full coverage. 
In more than half of these instances (13 cases), the RST annotation seems to  be inconsistent with the Strong Nuclearity principle\footnote{Note that difficulties in consistently annotating nuclearity have been pointed out before, see e.g., \cite{stede2008rst}}.  Figure \ref{fig:badalignment} illustrates this:
PDTB annotated a \rel{Synchrony} relation between segment (b) and (c). RST-DT annotated a \rel{Circumstance} relation between segment (a-b) and (c), but segment (a) is in fact annotated as the nucleus. In other words, the nucleus path could not be traced back to PDTB's Arg1, and therefore the relation was flagged. However, the \rel{Circumstance} relation does not actually hold between {\em The Kidder name is one of only six or seven} and {\em when considering a merger deal}; it holds between what PDTB marks as Arg1 and Arg2. 

Another typical case for automatic flagging by our algorithm occurred in cases where PDTB's annotation constraint for annotating only adjacent implicit relations caused a mismapping: this can happen when two adjacent sentences convey a similar message and are followed by a third sentence which is Arg2. In those cases, the second but not the first sentence has to be annotated as Arg1 in PDTB.


\section{Correspondence between mapped relation labels}\label{sec:mapping}
Our analysis of correspondences between mapped labels is based on a total of 5141 PDTB labels that could be mapped automatically with high confidence (see Section \ref{sec:align-results}). For any relation that carried two PDTB labels (in case the annotators thought that both relations held), we selected the label that is most similar to the corresponding RST-DT relation.

We will first discuss results concerning the discrepancies in theoretical mappings between frameworks as introduced in Section \ref{sec:theoretic}, and then analyse in more detail whether all mapped relations were annotated in a way that is consistent across frameworks, i.e. whether for a given relation, annotators from the two frameworks usually chose labels in PDTB that match the annotation in RST and vice versa. As results differ strongly between explicit and implicit relations, we will discuss this separately: explicit relations are discussed in Section \ref{sec:expl1} and implicit relations in Section \ref{sec:mapping-implicit}. 

\subsection{Analysis of theoretically interesting discrepancies in expected mappings}
\label{sec:map-theoretical}
Section \ref{sec:theoretic} laid out theoretically interesting discrepancies between proposed mappings, relating to highlighting the RST-DT relations \rel{Comparison}, \rel{Antithesis}, {\sc Elaboration-object-attribute}, \rel{Background} and \rel{Circumstance}. Table \ref{tab:theoreticCorrespondence} shows the mapping of relations corresponding to these labels.

\begin{table}[tbp]
\small
\centering
\caption{Alignment of RST-DT relations for those labels that were identified as theoretically interesting cases of discrepancies between the three mapping proposals. }
\setlength{\tabcolsep}{5pt}
\begin{tabular}{l|cccccccccccccc|c}
\hline
   & \multicolumn{2}{c}{Temp.} & \multicolumn{2}{c}{Cont.} & \multicolumn{3}{c}{Comp.} & \multicolumn{7}{c|}{Expansion} &     \\
RST-DT label  \hspace{1em} \sid{PDTB label}  & \sid{Synch.} & \multicolumn{1}{r|}{\sid{Asynch.}} & \sid{Cause} & \multicolumn{1}{r|}{\sid{Condition}} & \sid{Contrast} & \sid{Pragm. contr. } & \multicolumn{1}{r|}{\sid{Concession}} & \sid{Conjunction} & \sid{Chosen alt.} & \sid{Exception} &  \sid{Instantiation} & \sid{Specification} &\multicolumn{1}{r|}{\sid{List}} & \sid{EntRel}  & Total \\ \hline
Comparison            & 2            & 1     &  1  & 1 & 52       &                    & 2          & 24          &                    &           &   &  & 2      &  2  & 87    \\
Antithesis            & 1            & 3      &   &   & 186      & 2                  & 37         & 15          & 5                  & 2         &   1 &   &       &  2  & 254   \\
Elab.-object-attr. &           & 6     &     &    &      &                    &            & 7           &                    &           &     &  1  &   & 1   & 15   \\
Background        &           &   10    &  13   &    &   13   &                    &      2      &    19        &                    &           & 2    & 4 &       &   21  & 84  \\
Circumstance     &       90    &     83  &  39   & 22   &  15    &                    &     4       &     21       &                    &           &   &  5 &       &  15   &  294 \\ \hline
\end{tabular}
\label{tab:theoreticCorrespondence}
\end{table}

Regarding the first label, PDTB does not have a directly corresponding label for RST-DT's \rel{Comparison} label. 
Because the RST-DT annotation manual explicitly defines these relations as non-contrastive, \namecite{sanders2016} mapped this label to \rel{Conjunction} while \namecite{bunt2016} mapped it to \rel{Contrast}. 
Looking at the empirical results (see Table \ref{tab:theoreticCorrespondence}), we find that approximately one third of RST-DT \rel{Comparison} instances are annotated as \rel{Conjunction}, while two thirds (63\%) are annotated as \rel{Contrast} in PDTB; we specifically note a high proportion of PDTB \rel{Contrast.juxtaposition} labels for these instances. This distribution is similar for explicit and implicit instances of \rel{Comparison}. Typical markers for explicit instances are \textit{while}, \textit{but} and \textit{however}. We therefore conclude that RST-DT's {\sc Comparison} should be mapped to both PDTB {\sc Conjunction} and \rel{Contrast}.

A second area of disagreement was the label \rel{Antithesis}, which was proposed to map to \rel{Contrast} or \rel{Concession} in UniDim, but only to \rel{Concession} according to the ISO-based mapping and only to \rel{Contrast} according to OLiA. The empirical data clearly shows that the vast majority of \rel{Antithesis} relations (73\%) map onto PDTB's \rel{Contrast} relations, while 15\% map to \rel{Concession} relations, in line with OLiA and UniDim.


Next, we consider RST-DT's \rel{Elaboration-object-attribute}. OLiA proposed a correspondence with general \rel{Expansion} labels, UniDim proposed a mapping to \rel{Specification} and \rel{Generalization}, and ISO proposed a mapping to PDTB's \rel{EntRel}. The data only contained 15 instances of \rel{Elaboration-object-attribute}, and the majority of these (seven instances) were mapped to \rel{Conjunction}, confirming OLiA's mapping. Six cases were mapped to PDTB's \rel{Asynchronous} label, which none of the frameworks predicted. Only one instance was annotated as \rel{Specification}. The closest corresponding labels for \rel{Elaboration-object-attribute} therefore seem to be the more general \rel{Conjunction} label, as well as the temporal \rel{Asynchronous} type.

Finally, we consider the labels \rel{Background} and \rel{Circumstance}. UniDim mapped \rel{Background} to PDTB \rel{Conjunction} and \rel{Asynchronous}, whereas ISO only mapped it to \rel{Conjunction}. As shown in Table \ref{tab:theoreticCorrespondence}, \rel{Background} relations are annotated as a variety of relations:  23\% are annotated as PDTB \rel{Conjunction}, and 12\% as \rel{Asynchronous}, but \rel{Cause} and \rel{Contrast} (both 15\%) also occur frequently. 

Regarding \rel{Circumstance}, UniDim mapped it to PDTB \rel{Conjunction}, \rel{Synchronous} and \rel{Asynchronous}, and ISO only mapped it to \rel{Synchronous}. Table \ref{tab:theoreticCorrespondence} shows that \rel{Circumstance} is also annotated as a wide range of PDTB relations, indeed including \rel{Conjunction} (7\%), \rel{Synchronous} (31\%) and \rel{Asynchronous} (28\%), but also \rel{Cause} (13\%) and \rel{Condition} (7\%).
RST-DT's \rel{Background} and \rel{Circumstance} therefore seem to be more general labels than both UniDim and ISO predicted. This could be due to the lack of corresponding label in PDTB.
\smallskip

We conclude that the empirical data confirms the usage of the label \rel{Antithesis} primarily for contrast relations rather than concessives; we would suggest that any new annotation efforts using the \rel{Antithesis} label should clarify its intended use in the annotation guidelines; the same holds for the label \rel{Comparison}, which would need to be more clearly defined for future use to avoid misinterpretation. The mapping of relations \rel{Background} and \rel{Circumstance} to PDTB is difficult, because these labels are more general, used when additional information needs to be provided to allow the reader/listener to understand. Therefore, these segments have a function within the overall goal of the text, but do not seem to stand in a consistent semantic relation to the segment that they are supposed to provide additional context for. 
We now turn to the findings for the mapping of other labels to see if the two frameworks' annotations are compatible.

\subsection{Mapping for explicitly marked relations}
\label{sec:expl1}
Table \ref{expl} displays the mapping of PDTB annotations onto RST-DT annotations for explicitly marked discourse relations that occurred more than 30 times in total. In the table, relation mappings which were suggested by all three proposals are considered ``expected'' mappings and are marked in the table by underlined, bold numbers. Mappings of labels for which at least two of three proposals agree are indicated by underlined numbers.
Note that for some labels, several entries in a row or column are marked as ``expected''. This is because a one-to-one mapping is not always possible, as some labels have multiple matching candidate labels in another framework due to differences in the granularity of distinctions between frameworks (see Section \ref{sec:theoretic}). 
We can distinguish a number of clusters that correspond quite well to the expected mappings. In this section, we will go through the major classes (temporals, causals, contrastives, additives).


\begin{table}[t]
\small
\centering
\caption{Alignment of explicit discourse relation classes for which $n>$30. Numbers indicate how many instances occurred in our high-confidence mapping; values are represented as colors. Underlined, bold numbers indicate the mapping predicted to all three proposals; numbers that are only underlined (not bold) are predicted by two of three proposals. }
\begin{tabular}{l|ccccccccc|c}
\hline
   & \multicolumn{2}{c}{Temp.} & \multicolumn{2}{c}{Cont.} & \multicolumn{2}{c}{Comp.} & \multicolumn{3}{c|}{Expansion} &     \\
RST-DT label    \hspace{2.7em} \sid{PDTB label}     & \sid{Synch.}  & \multicolumn{1}{r|}{\sid{Asynch.}}  & \sid{Cause}        & \multicolumn{1}{r|}{\sid{Condition} }       & \sid{Contrast}      & \multicolumn{1}{r|}{\sid{Concession}}     & \sid{Conjunction} & \sid{Instantiation} & \sid{List} & Total \\ \hline
Temp.-same-time        & \cellcolor{red7}{\expect{63}}               & 1                    & 1                 & 1                     & 4                   &                      & 5                     &                        &               & 75   \\
Temporal-after            &                   & \cellcolor{red8}{\expect{48}}                   &                  &                      & 1                   &                      & 2                     &                        &               & 54   \\
Sequence                  & 2                  & \cellcolor{red8}{\expect{29}}                   & 1                 &                      & 1                   &                      & {19}                    &                        &               & 52   \\
Circumstance              & \cellcolor{red6}{\underline{89}}  & \cellcolor{red6}{79}                   & \cellcolor{red8}{29}             & {21}                    & 7                   & 4                     & {10}                    &                        &               & 259  \\
Result                    & 1                  & 3                    & \cellcolor{red8}{\expect{30}}                &                      & 2                   &                      & {5}                     &                        &               & 41   \\
Consequence             & 5                  & 6                    & \cellcolor{red8}{\expect{45}}                &                      & 4                   & 1                     & {16}                     &                        & 1              & 95   \\
Explanation-arg. & {6}                  &                     & \cellcolor{red8}{\expect{39}}                &                      & {7}                   & 2                     & 1                     &                        &               & 57   \\
Reason                    &                   & 2                    & \cellcolor{red7}{\expect{67}}                &                      &                    &                      &                      &                        &               & 72   \\
Condition                 & 5                  & {13}                   &    1              & \cellcolor{red5}{\expect{104}}                   & 2                   & 1                     & 1                     &                        &               & 182  \\
Contrast                  &   1                & 1                    &                  &                      & \cellcolor{red4}{\expect{160}}                 & {23}                    & {17}                    &                        &               & 208  \\
Concession                & 5                  & 4                    &                  & 3                     & \cellcolor{red5}{101}                  & \cellcolor{red7}{\expect{53}}                    & 4                     &                        &               & 182  \\
Antithesis                & 1                  & 3                    &                  &                      & \cellcolor{red4}{\underline{170}}                 & \cellcolor{red8}{\underline{37}}                    & 10                    &                        &               & 243  \\
Comparison                & 2                  & 1                    &                  &                      & \cellcolor{red8}{26}                  & 2                     & {\expect{9}}                     &                        &               & 41   \\
Elaboration-add.    & 4                  & 3                    & 1                 &                      & \cellcolor{red8}{30}                  & {8}                     & \cellcolor{red5}{\underline{122}}                   & 3                       & {3}              & 192  \\
Example                   & 1                  &                     &                  &                      & 1                   &                      & {3}                     & \cellcolor{red8}{\expect{29}}                      &               & 35   \\
List                      & 2                  & 4                   & 1                 &                      & 17                  & 1                     & \cellcolor{red1}{303}                   &                        & \cellcolor{red8}{\expect{47}}             & 377  \\ \hline
Total                          & 187                & 197                  & 215               & 129                   & 533                 & 132                   & 527                   & 32                      & 51             &  \\
\end{tabular}
\label{expl}
\end{table}

\paragraph{Temporals} 
The results show that most (81\%) of the explicitly marked relations that were classified as \rel{Synchronous} by PDTB were tagged as RST-DT \rel{Temporal-same-time} or \rel{Circumstance}. There are however also some cases where annotations deviated from expected mappings for temporal PDTB labels. These include cases where one of RST-DT's causal labels (specifically, \rel{Explanation-argumentative} or \rel{Consequence}) was annotated. Closer inspection revealed that frequent connectives in these relations, which did not receive a temporal sense label in RST-DT, were \textit{as} and \textit{when}. These connectives are known to be ambiguous markers~\cite[see e.g.~][]{asr2013}, and are frequent among temporal relations such as \rel{Circumstance} and \rel{Temporal-same-time} in RST-DT. We hence find that there are some instances containing these ambiguous connectives which could not be consistently disambiguated between frameworks, with one framework labelling these instances as temporal and the other as causal. We will analyse the annotation of ambiguous connectives in more detail in Section \ref{sec:connectives}

PDTB's temporal \rel{Asynchronous} relations generally also map well to their corresponding RST-DT classes (79\%). The most notable unexpected pattern consists of PDTB temporal relations marked with \textit{until} often being classified as \textit{Condition} in RST. This mismatch could be indicative of inconsistencies in disambiguation of this marker, or could be more systematically related to RST-DT annotating the intention in subjective relations, while PDTB annotations stay closer to the semantic relation \cite{scholmansubm}.

For RST-DT's \rel{Sequence} class, we find that a substantial portion is annotated as PDTB's \rel{Conjunction} (37\%), i.e.~annotations by the different frameworks disagree in these instances as to whether the relation is temporal or not. This was not predicted by any of the mapping proposals.


\paragraph{Causals}
Explicit causal and conditional PDTB relation labels generally map well onto causal and conditional RST-DT labels; a finding that can be attributed to the relative definiteness of causal markers (we will see a different results for implicit relations in Section \ref{sec:mapping-implicit}). RST-DT distinguishes more types of causal relations, which results in causal PDTB relations being distributed among the various causal RST-DT classes. Unexpected mappings -- e.g., PDTB causals and conditionals annotated as RST-DT's \rel{Circumstance} (12\% and 16\%, respectively) -- were found to occur again for instances marked by the ambiguous connectives \textit{as} and \textit{when}, respectively.

\paragraph{Contrastives}
The majority of PDTB's \rel{Contrast} relations was mapped to RST-DT's \rel{Contrast} and \rel{Antithesis} relations (62\%), as expected. However, we also found that a substantial portion (19\%) of PDTB's \rel{Contrast} relations are annotated as RST-DT's \rel{Concession}; and some also as RST-DT's \rel{Elaboration-additional} (6\%). 
These cases  were often marked by the connective \conn{but}, which is an ambiguous connective.

A closer look at the subtypes of PDTB \rel{Concession} reveals that relations annotated as PDTB's \rel{Concession.Expectation} map quite well (54\%) onto RST-DT's \rel{Concession} relations, while \rel{Concession.Contra-expectation} relations are often annotated as \rel{Contrast} in RST, especially when marked with the connective \textit{but}. We note that the distinction between concession and contrast relations is known to be difficult in discourse relation annotation \cite{robaldo2014}. It is possible that the observed differences stem from differences in interpretation between annotators, and slight biases in the frameworks.
Overall, PDTB has a stronger bias than RST-DT towards assigning the \rel{Contrast} label: the majority of three RST-DT relational labels, namely \rel{Contrast}, \rel{Antithesis} and \rel{Concession}, are mapped to PDTB's \rel{Contrast}. 

\paragraph{Additives}
Finally, looking at PDTB's \rel{Expansion} relations, we find that a majority of relations annotated as PDTB's \rel{Conjunction} is annotated as RST-DT's \rel{List} (57\%), which was not expected based on the theoretical definitions of these relations. Closer inspection shows that the high number of cases annotated as RST-DT \rel{List} stems from the fact that PDTB annotation guidelines say that lists have to be ``announced''. Unannounced lists cannot be annotated as \rel{List} relations in PDTB. Such a criterion is however not applied in RST-DT.
We find that PDTB's \rel{List} and \rel{Instantiation} relations map well onto RST-DT's \rel{List} and \rel{Example} relations respectively. 

We also observe a substantial amount of noise, i.e.~26\% of \rel{Conjunction} relations have a temporal, causal or contrastive label in RST; some of these cases contain the connectives \textit{but} or \textit{while}, again indicating that connective disambiguation may not always be consistent between frameworks.

A final interesting observation regards the annotation of the connective \textit{unless}: these instances are annotated as \rel{Condition} in RST-DT but as \rel{Alternative.Disjunctive} relations in PDTB (note that RST-DT does not have a corresponding label). 
\bigskip

In sum, we find that deviations from expected mappings are often related to ambiguous connectives (e.g.,~\textit{as}, \textit{when}, \textit{but}, \textit{while}) which are in some instances resolved differently in the two frameworks, as well as differences in operationalization between frameworks which leads to mismatches in annotations for relations such as \rel{list}. We will now consider the annotation of connectives in more detail, before turning to the mapping of implicit relations.

\subsubsection{Analysis of annotation for ambiguous connectives}\label{sec:connectives}
Given that the disagreements between the PDTB and RST-DT can be related to different interpretations of specific ambiguous connectives (as discussed in Section \ref{sec:expl1}), we studied the agreement on annotating ambiguous connectives in more detail. After all, it is not surprising that annotators can reach high agreement on relations with an unambiguous connective such as \textit{if} for \rel{Condition}; given the empirical findings from previous annotation work in English, this could be done automatically in the future. Rather, the value of additional manual annotation comes from disambiguating between relations when the connective can mark different relations (or when a relation is not explicitly marked, see Section \ref{sec:mapping-implicit}).

\paragraph{Methods}
To investigate the agreement on connectives, we tested the independence of PDTB and RST-DT annotations by calculating a separate $\chi^2$ test for each connective. For connectives where the distribution violated the assumptions of the $\chi^2$ test (less than 5 expected observations in a cell), we instead used the non-parametric  Fisher's exact test. 
These analyses reveal whether the annotations agreed more with each other than can be expected based on the distribution of the connectives in the data.
The results will provide insight into whether the content of the discourse relation arguments had been taken into account for the actual annotations of the relation instances, or whether the distinction is potentially too difficult or too subtle for human annotators to make reliably.
Note that this analysis is more strict than the usual kappa for inter-annotator agreement, because we use the distribution of relations per connective (i.e.~which relations a connective can mark, and how often it does so for each relation type in the text at hand), which we are not normally known.

\paragraph{Results}
We report some representative results for connectives where the null hypothesis of PDTB and RST-DT annotations for a connective being independent from one another. Here, independence would mean that the content of the actual content of an instance is not predictive of which label was chosen by the one vs.~the other framework; in the ideal case, we would expect strong non-independence: given a pair of segments, the labels chosen by one and the other framework should correspond to one another and hence be deterministic not random. We first report cases for which the labels corresponded well (i.e.~the null hypothesis of labels being independent could be rejected with high confidence), and then discuss connectives for which observed distributions were very similar to random distributions.

The connective \textit{while} is an example of the first case: we find that annotators could distinguish between the \rel{Temporal.Synchronous} vs.~\rel{Contrast} / \rel{Comparison} reading of \conn{while} very well; the annotations from the two frameworks almost always agreed on the reading ($p<0.0001$).

On the other hand, the distribution for sense labels of connectives which are ambiguous between more similar discourse relations (\rel{Contrast} vs.~\rel{Concession}), such as \textit{but}, \textit{although} and \textit{however}, did not significantly differ from a random distribution of these sense labels (given the marginals), according to a $\chi^2$ test for \textit{but} and Fisher's exact tests for \textit{although} and \textit{however}. Calculating $\kappa$ values in this strict reading (i.e.~taking for granted that \textit{but} cannot mark causals or temporals and only testing agreement on different subtypes of negative relations) corresponds to $\kappa$ $< 0.1$ for these relation label distinctions between frameworks. 

To summarize, we find that there are some ambiguous connectives for which manual annotation from the two frameworks reliably agreed on how the connective should be disambiguated and hence provide valuable additional information. This was mostly the case for when the alternative readings of the connective strongly differ from one another. However, we found no conclusive evidence that most subtle distinctions could be made reliably -- on these cases, annotations from the two frameworks often don't agree with one another more than would be expected by random assignment of labels that a connective can occur with (given the distribution of these connectives). This lack of agreement between humans may also provide a partial explanation for why automatic discourse relation disambiguation is difficult -- it is unclear whether the training data that these distinctions are trained on is fully consistent internally, and/or the distinction may be so subtle in a substantial number of real data cases that even humans find it hard to agree. We will next move on to the analysis of agreement between frameworks on implicit discourse relation annotation.

\subsection{Mapping of implicit discourse relations}
\label{sec:mapping-implicit}
The overall picture of annotation agreement between the two frameworks looks a lot more problematic for implicit than for explicit relations. To get an idea of what underlying causes these differences stem from, we decided to provide two perspectives on the results: once from the PDTB view, which provides an overview of how RST labels are distributed for a given PDTB relation (Table \ref{implPDTB}) and once from the RST view, showing how the PDTB labels are distributed for a given RST relation (Table \ref{implRST}). 
To keep tables readable, we only include those labels that occurred more than twenty times in the data. 

Mapping of relation annotations for implicit relations, as seen from the PDTB perspective, is shown in Table \ref{implPDTB}. Expected correspondences in the table are again indicated by underlined numbers.
The colours in the table indicate the percentage of correspondence to RST-DT labels, with darker shades indicating a higher proportion of instances with a certain PDTB label falling into that RST-DT category. For example, 7 instances of PDTB \rel{Temporal.Asynchronous} relations are annotated as RST-DT \rel{Background}. As these 7 instances represent more than 10\% of the data on \rel{Temporal Asynchronous} relations, the number is shaded in light green. On the other hand, the 7 counts of \rel{Comparison.Contrast} relations labelled as RST's \rel{Consequence} represent less than 10\% of the 219 PDTB \rel{Comparison.Contrast} relations; the entry is therefore not shaded. 

A first striking observation is that the agreement between frameworks is a lot worse than for explicit relations. While we saw a diagonal line of shaded cells that largely overlapped with theoretically expected correspondences, we can see that a substantial proportion of instances from almost all PDTB classes were annotated as RST-DT's \rel{Elaboration-additional}. 

Table \ref{implRST} shows the alignment of implicit relations from the RST-DT perspective. Many cells here are shaded green, which indicates that the annotations for many of the RST-DT relations have a wide variety of PDTB labels.

\paragraph{Temporals}
Temporal relations were not consistently identified between frameworks. Only roughly one third of PDTB's \rel{Asynchronous.Precedence} relations had the expected RST label \rel{Sequence}, and most of PDTB's \rel{Asynchronous.Succession} relations were labelled as \rel{Elaboration-additional} in RST-DT. The RST-DT \rel{Temporal-after} label is very rarely annotated among implicit relations.

\paragraph{Causals}
For PDTB's \rel{Cause.Reason} relations, less than 40\% of instances were annotated as one of the expected causal classes by RST-DT annotators (expected classes were \rel{Explanation-argumentative}, \rel{Reason}, \rel{Evidence} or \rel{Interpretation}), and only a very small percentage of PDTB's \rel{Cause.Result} were annotated as \rel{Consequence}, \rel{Result} or \rel{Cause-result} relations in RST. Instead, most of these relations are annotated as \rel{Elaboration.additional}, showing that PDTB annotators tended to choose a ``stronger'' label than RST annotators for these instances.

RST-DT's causal relations \rel{Consequence} and \rel{Reason} map relatively well onto PDTB's causal relations. However, other causal RST-DT labels (\rel{Evidence} and \rel{Explanation-argumentative}) are often mapped onto the additive PDTB labels \rel{Instantiation} and \rel{Specification}. This difference can be attributed to a fundamental difference between the approaches: PDTB annotates the lower-level ideational relations between arguments, while RST-DT focuses more on the intentional level. \namecite{scholmansubm} show that often two functions can be identified in these specific relations: a segment can provide an example or a specification of a set, as well as providing evidence for a previously stated claim. These double functions are reflected in the annotation mapping between PDTB and RST.

\paragraph{Contrastives}
Only a minority of PDTB \rel{Contrast} relations was also labelled as a contrastive relation in RST (17\% for RST \rel{Contrast}, 10\% for \rel{Comparison}). Instead, the majority is mapped to \rel{Elaboration-additional} and even \rel{List} labels. 

From the perspective of RST-DT's \rel{Contrast} relations, we observe that they were for the most part annotated as contrastive relations in PDTB, usually using the underspecified \rel{Contrast} label (rather than one of its subtypes). 


\paragraph{Additives} 
One PDTB implicit relation that matches well with RST-DT annotation is \rel{List}, for which 84\% of instances were annotated as RST-DT's \rel{List} relation. For RST-DT's \rel{List} relation, we see that the largest proportion of its instances (44\%) are annotated as \rel{Conjunction} in PDTB; as mentioned earlier, this problem is partially due to the guideline in PDTB that lists have to be announced. 

Taking a look at PDTB \rel{EntRel} labels, we can see that these are predominantly annotated as \rel{Elaboration-additional} and \rel{List} in RST-DT. Nevertheless, we also observe a number of different annotations on the RST side. We analysed a randomly selected subset of these instances and found that a common reason for the richer RST labels is that these EntRel relations are in fact high-level relations in RST. The stronger interpretation is in those cases often due to the effect of additional content (outside the nucleus path), which facilitates the stronger interpretations. In our view, it is arguable whether EntRel labels should really be considered to be mapped to RST labels, as the task given to the annotators differed markedly for these instances: in the PDTB case, the annotator task was to label the relation holding between two adjacent sentences, whereas the task of the RST annotators was to join high-level discourse segments and describe their relation. 

RST-DT's  \rel{Comment} relation shows an almost uniform distribution across PDTB labels, consistent with other communicative functions which are not represented in the PDTB annotation scheme such as \rel{Background} and \rel{Circumstance}, which have been discussed in Section \ref{sec:map-theoretical}. 

Finally, we observe that RST-DT's \rel{Elaboration-general-specific} relations are often labelled as PDTB's \rel{Restatement} (55\%). This correspondence was predicted according to the expected mappings. 21\% of instances were labelled as \rel{Instantiation}, which can be attributed to the subtlety of the distinction between these two labels.

\begin{table}[h!]
\footnotesize
\caption{Alignment of implicit discourse relation classes for which $n>$20. Numbers indicate how many instances occurred in our high-confidence mapping. }
\begin{subtable}{.99\textwidth}
\caption{Colours encode percentage agreement from PDTB perspective, i.e. darker colours show that most instances of a PDTB relation  type occurred in that specific RST-DT class.}
\begin{tabular}{l|ccccccccc}
\hline 
 \hspace{-0.9em}  \includegraphics[width=2.6cm]{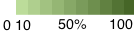}  
   & Temp. & Cont. & Comp. & \multicolumn{4}{c}{Expansion}   &    \\
RST-DT label    \hspace{2em} \sid{PDTB label}   & \multicolumn{1}{c|}{\sid{Asynch.}}  & \multicolumn{1}{c|}{\sid{Cause}} & \multicolumn{1}{c|}{\sid{Contrast}} & \sid{Conjunction} & \sid{Alternative} & \sid{Instantiation} & \sid{Restatement} & \sid{List}  &  \multicolumn{1}{|c}{\sid{EntRel}} \\ \hline
Sequence                  &      \cellcolor{heat7}{\expect{16}}               &    1               &     2             &        4           &           &     &               &                     & 3  \\
Background             & {7}   & 12  & 10       & {16}  &   & 2           &  4      &      &     21     \\
Circumstance          & \expect{1}     & 10                & 8                   & {11} &   &                        &    6           &                &  15     \\
Consequence          &  {4}                   & \expect{16}  & 7                   & 11  &                  & 2                    & 2            & 1              &    4    \\
Evidence                 &                      & \expect{12}   & 2                   & 12            &   2     & \cellcolor{heat8}{29} &     {28}  &            &       6    \\
Explanation-arg.     &        2             & \cellcolor{heat7}{\expect{114}} &  16      &      16   &  2                 & \cellcolor{heat8}{35}                      & \cellcolor{heat8}{51}                  &  1          &    22   \\
Reason                  &     1                & \expect{20}                    & 1                & 3          &        & 1                     & 2                  &            &    3   \\
Result                  &       1              &        \expect{12}           &      1            &           4        &           &     &       3        &         1            &  5 \\
Evaluation              &                     & 7                                     & 9               & 11          &        & 2                     & 1               &           &    12   \\
Interpretation          &                       & \expect{16}   & 3                   & 9          &           & 2                    &    8           &                    &  8  \\
Contrast                  &        1             & 2                 & \cellcolor{heat8}{\expect{35}}     & 2    &               &                &               & 2                    &  3 \\
Comparison            &                      & 1                 & {24}                  & 14  &       \cellcolor{heat8}{4}         &                 &               &  2               &   2  \\
Antithesis                  &                     &                   &           \underline{15}       &           5        &   2        &   1  &               &                     & 2  \\
Elaboration-add.      & \cellcolor{heat6}{29}  & \cellcolor{heat5}{168} & \cellcolor{heat6}{73}  & \cellcolor{heat5}{\underline{221}} & \cellcolor{heat6}{10} & \cellcolor{heat8}{36}    &  \cellcolor{heat5}{151}     & {7}     &    \cellcolor{heat3}{266}  \\
Example                  &                   & 9                 & 1                   & 6                     &    & \cellcolor{heat6}{\expect{64}}                     &      21          &          &    2   \\
Elab.-gen.-spec.     &                     & 8                  &                    & 11                 &         &  {17}              &   \cellcolor{heat8}{\expect{44}}            &  &  18 \\
List                          & {6}             & 24               & \cellcolor{heat8}{29} & \cellcolor{heat7}{120}  & 2 & 6                       &  13        & \cellcolor{heat1}{\expect{74}}               & {30}  \\
Restatement                  &                     &        5           &                  &      2             &  1         &   2  &   \expect{12}            &                     & 1  \\
Comment               &                     & 16                & 9                 & {8}     &        &                    &   {4}                &          &  8   \\ \hline
Total                       & 75                 & 499               & 277             & 511        &      26       & 202              &   406              &   88     &   463 \\ \hline
\end{tabular}
\label{implPDTB}
\end{subtable}
\smallskip

\begin{subtable}{.99\textwidth}

\caption{Colours encode percentage agreement from RST-DT perspective, i.e. darker colours show that most instances of a RST-DT relation  type occurred in that specific PDTB class.}
\begin{tabular}{l|ccccccccc|c}
\hline
   & Temp. & Cont. & Comp. & \multicolumn{4}{c}{Expansion} &   &  \\
RST-DT label    \hspace{2em} \sid{PDTB label}   & \multicolumn{1}{c|}{\sid{Asynch.}}  & \multicolumn{1}{c|}{\sid{Cause}} & \multicolumn{1}{c|}{\sid{Contrast}} & \sid{Conjunction} & \sid{Alternative} & \sid{Instantiation} & \sid{Restatement} & \sid{List} & \multicolumn{1}{|c|}{\sid{EntRel}} &  Total   \\ \hline
Sequence                  &          \cellcolor{heat4}{\expect{16}}           &     1              &    2              &      \cellcolor{heat8}{4}             &           &     &               &     &     3           & 27  \\
Background                 & {7}  & \cellcolor{heat8}{12}   & \cellcolor{heat8}{10}    & \cellcolor{heat7}{16}          &          & 2                       &      4         &                   & \cellcolor{heat7}{21}  & 72   \\
Circumstance              & \expect{1}    & \cellcolor{heat7}{10}                & \cellcolor{heat8}{8}                   & \cellcolor{heat7}{11}         &           &                        &      \cellcolor{heat8}{6}       &         &  \cellcolor{heat7}{15}    & 51  \\
Consequence             &  {4}   & \cellcolor{heat6}{\expect{16}}                & \cellcolor{heat8}{7}                   & \cellcolor{heat7}{11}         &           & 2                       & 2             & 1                  & {4}  & 49   \\
Evidence                     &                     & \cellcolor{heat8}{\expect{12}}      & 2    & \cellcolor{heat8}{12}     &  2    & \cellcolor{heat6}{29}              &         \cellcolor{heat6}{28}      &      &  {6} & 92   \\
Explanation-arg.         &      2               & \cellcolor{heat5}{\expect{114}}    & 16                  & 16           &         & \cellcolor{heat7}{35}       & \cellcolor{heat7}{51}            &  1   & {22}  & 267  \\
Reason                  &     1                  & \cellcolor{heat3}{\expect{20}}                    & 1                & {3}    &              & 1                     & 2                  &            &    {3} &  31 \\
Result                  &      1              &     \cellcolor{heat5}{\expect{12}}              &      1            &       \cellcolor{heat7}{4}            &           &     &        3       &        1             & \cellcolor{heat8}{5} & 28 \\
Evaluation              &                          & \cellcolor{heat8}{7}    & \cellcolor{heat7}{9}               & \cellcolor{heat7}{11}      &            & 2                     & 1               &           &    \cellcolor{heat7}{12}  & 46 \\
Interpretation              &                     &  \cellcolor{heat5}{\expect{16}}      & 3    &  \cellcolor{heat7}{9}  &   & 2                       &           \cellcolor{heat7}{8}          &              &  \cellcolor{heat8}{8} & 49   \\
Contrast                     & 1                    & 2                 &  \cellcolor{heat1}{\expect{35}}                  & 2           &          &                        &              &     2            &   3  & 56   \\
Comparison               &                     & 1                 &  \cellcolor{heat4}{24}      &  \cellcolor{heat6}{14}    &    4            &                        &              &  2                 & 2  & 44   \\
Antithesis                  &                     &                   &         \cellcolor{heat4}{\underline{15}}         &           \cellcolor{heat8}{5}        &          2      &      1         &       &      &   2     & 26   \\
Elaboration-add.        & 29                  &  \cellcolor{heat8}{168}     &  {73}                  &  \cellcolor{heat7}{\underline{221}}   & 10  & 36                      & \cellcolor{heat8}{151}              &  7              &  \cellcolor{heat7}{266}   & 991  \\
Example                    &       1              & {9}                 & 1                    & 6               &      &  \cellcolor{heat3}{\expect{64}}                      &   \cellcolor{heat7}{21}                &              &  2  & 106  \\
Elab.-gen.-spec.        &                     &  {8}                 &                    &  {11}         &           &  \cellcolor{heat8}{17}                      &       \cellcolor{heat5}{\expect{44}}          &               &  \cellcolor{heat8}{18}   & 99   \\
List                            & 6                    &  {24}                &  {29}                  &  \cellcolor{heat5}{120}     &    2          & 6                       &   13         &  \cellcolor{heat7}{\expect{74}}                 &  {30}   & 311   \\
Restatement                  &                     &      \cellcolor{heat8}{5}             &                  &       2            &    1       &    2 &    \cellcolor{heat5}{\expect{12}}           &                     & 1 & 25 \\
Comment                   &                     &  \cellcolor{heat5}{16}                &  \cellcolor{heat8}{9}                   &  \cellcolor{heat8}{8}   &                  &                        &     {4}              &              &  \cellcolor{heat8}{8}   & 49   \\ \hline
\end{tabular}
\label{implRST}
\end{subtable}
\vspace{-.3cm}

\end{table}

\paragraph{Discussion on correspondence of annotations for implicit relations}
Our analysis of the implicit relations from both perspectives documents a low level of agreement between labels assigned by RST-DT and PDTB. 
Generally, a stronger relation (biasing away from annotating simple additive labels) tended to be chosen by PDTB annotators. Nevertheless, wherever the RST-DT annotators chose a label other than \rel{Elaboration-additional} (the predominant label assigned to the implicit cases), the annotations matched with their PDTB equivalents relatively well. 

These results raise the question of how these very substantial differences in annotations of implicit relations can be explained. We think that the discrepancy can be attributed largely to the differences in annotation guidelines and operationalizations for implicit discourse relations. PDTB's connective-driven approach biases against annotating simple additive relations when a connective can be inserted and hence an additional stronger interpretation of the discourse relation is available. RST-DT prescribes a different strategy: annotators are asked to annotate the writer's intentions. The resulting low agreement between PDTB and RST-DT on implicit relations have implications for the reliability and validity of these annotations. We will expand on this point in the Discussion.

\section{Related work}\label{sec:related}

The only other systematic evaluation of the mapping of discourse annotations from different frameworks on the same text is the work by \namecite{rehbein2016}. Rehbein et al.~created an English corpus of spoken discourse containing PDTB 3.0 and CCR annotations for every relation. They segmented the texts first, and then proceeded to assign sense labels according to both schemes for that given segmentation. This procedure thus avoided challenges related to differences in segmentation between frameworks.
After annotation, the relation labels were mapped onto one another directly, and the correspondence between PDTB and CCR annotations was evaluated. 

\namecite{rehbein2016} reported three systematic biases introduced in the operationalizations of PDTB 3.0 and CCR, which lead to differences in annotations in some areas: A first observation holds that PDTB's additive relations \rel{Expansion.Instantiation}, \rel{Expansion.Restatement.Specification} and \rel{Expansion.Restatement.Equivalence} were quite often (30\% of relations) annotated as causals in CCR. This finding is consistent with our observation here, where we found that PDTB \rel{Expansion.Instantiation} and \rel{Expansion.Restatement} are often annotated with a causal label in RST.
As noted by \namecite{blakemore1997} and \namecite{carston1993}, and shown in \namecite{scholmansubm}, these types of discourse relations are often ambiguous, as examples can at the same time also serve as evidence for a claim.

The second category of systematic disagreements concerns {\sc Comparison.Contrast} and {\sc Comparison.Concession} relations: among the negative relations, annotators often disagreed on the causal vs.~additive basic operation. This was partly due to a slightly different definition of what constitutes a {\sc Concession}, but note that distinguishing between contrastive and concessive discourse relations is a well-attested difficulty \cite[see, for example,][]{robaldo2014,zufferey2013}. Again, the same difficulty is obvious in the mapping between RST-DT and PDTB, as discussed in Section \ref{sec:map-theoretical}.

As a third pattern of disagreements, \namecite{rehbein2016} report effects of operationalization of annotation procedures: Some instances marked by \conn{but} were annotated as positive polarity relations in PDTB, but as negative in CCR (including instances marked with \textit{but also}). These discrepancies were systematic and due to an annotation instruction -- as a rule,  all relations that can be marked with \conn{but} are annotated as negative polarity relations in CCR. While this specific pattern is not relevant for the mapping between PDTB and RST-DT, we have nevertheless also observed substantial effects of annotation operationalization here, such as the large discrepancies in terms of annotating implicit discourse relations, with PDTB preferring more specific labels due to the annotation instruction to in a first step try to insert a connective and then annotate the relation. 

%
%

\section{Discussion}\label{sec:discussion}

In the current paper, we evaluated how well theoretical proposals for mapping discourse relational labels correspond to the mapping between existing RST-DT and PDTB 2.0 annotations. The study consisted of three main steps: alignment of relations, evaluation of theoretically interesting discrepancies between the proposals, and evaluation of the mapped labels for explicit and implicit relations. We will discuss each one in turn. 

\paragraph{Automatic alignment} In order to evaluate existing annotations, we proposed an automatic mapping algorithm for PDTB and RST-DT discourse relation annotations and applied it to a segment of the WSJ corpus that contains annotations from both frameworks. 
The algorithm proposed in this article allowed us to align 76\% of PDTB discourse relation annotations to corresponding RST annotations. 
Our manual error analysis shows that the alignment algorithm is highly accurate; it also correctly identifies instances where annotations cannot be aligned due to more fundamental differences in how the discourse is analysed. 

Our study highlights the importance of discourse segmentation. Segmentation has a strong effect on determining the scope and argument structure of a discourse relation \cite[see also][]{hoek2017}. The differences in segmentation may hold interesting insights about effects of operationalization of discourse segmentation on discourse annotation, which could be explored in future work to refine annotation processes both for manual annotation and automatic processing.

\paragraph{Evaluation of mapping proposals} As a result of the annotation alignment, we are able to offer a more complete picture of how annotations from the two frameworks relate to one another in practice. We compared actual annotations to expected correspondences which were determined based on three recent proposals for mapping discourse relations onto one another. 
Our aim in evaluating three different mapping proposals was not to identify the ``best'' proposal; rather, 
the comparison between proposed correspondences and empirical co-ocurrences of annotated labels is helpful for achieving a deeper understanding of how certain definitions in the annotation guidelines were applied in practical annotation, and can help to decide between alternative proposals for mappings. The observed mismatches between the proposals can furthermore be used for clarifying annotation guidelines in future annotation projects; for example, the three proposals differed in their interpretation of RST-DT's \rel{Antithesis} label, which indicates that the definition of this label could be expanded on.

We found high numbers of disagreement between observed and expected annotations for those relations that did not have a direct correspondence in the other scheme. Examples of such relations include RST's \rel{Background} and \rel{Circumstance} relations. The three proposals treated these relations differently from each other based on the definitions in the annotation manual, but generally, they were mapped to temporal or additive labels. However, the empirical mapping showed that many of these relations were also annotated as causals or even contrastives in PDTB. We see two possible explanations for what causes this: either the mapping scheme (and possibly the annotation manuals) would have to be revised to more clearly or exhaustively describe the relations, or there is a function to the relation which is not reflected in the PDTB scheme, and should be considered to be added to relation schemes, again possibly by annotating both the ideational and intentional functions separately.

\paragraph{Evaluation of mapped explicit and implicit relations}
After evaluating the theoretical discrepancies between the proposals, we looked at the mapped data for explicit and implicit relations separately.
The most striking observations were a lower than expected level of agreement on annotations for implicit relations, and low agreement on more fine-grained distinctions for explicitly marked relations. 
In order to get more insight into this issue of difference in agreement, we recommend that future annotation efforts (corpus annotation as well as other tasks) report agreement on implicit and explicit relations separately.
While the rate of non-correspondences does seem problematic, we were able to identify several patterns that lead to these observed disagreements. Specifically, many of the differences can be traced back to different operationalizations employed during annotation. as well as to the different goals of PDTB 2.0 vs. RST-DT annotation.  

First, the operationalization of discourse annotations may have a strong effect on the resulting annotations: in the PDTB annotation process, annotators were asked to annotate implicit relations by first identifying a discourse connective that would fit the relation, and then in a second step annotate the relation sense. It seems that this practice encourages annotators to assign more specific relation labels to implicit relations than RST-DT's annotation procedure does. We therefore find that most implicit relations receive the RST-DT label \rel{Elaboration-additional} and a more specific PDTB label. For future annotation efforts, it is important that this consequence of annotation operationalization is taken into account. We cannot determine whether RST-DT annotators relied too heavily on the absence of a connective, which may have biased them to annotate the \rel{Elaboration-additional} relation more often, or whether the insertion of connectives as a task made some of the interpretations of relations stronger than they were without the connective, i.e.~whether the operationalization to insert a connective may have changed the inferred relation in some cases. To answer this question, we suggest systematic annotation experiments for measuring effect of annotation instructions. 

Second, the framework-specific goals have led to a focus on different levels of analysis of discourse relations, namely the \textit{ideational} and the \textit{intentional} level. Ideational relations describe the semantic relation between the information conveyed in the consecutive elements of a coherent discourse \cite[cf.][]{moore1992}. \textit{Intentional relations} on the other hand involve the writer's attempts to affect the addressee's beliefs, attitudes, desires etc.~by means of language (cf. \citealt{hovy1995}; see also \citealt{crible2017,redeker1990}).\footnote{The ideational and intentional levels are also referred to as informational vs.~intentional \citep{moore1992}, subject matter vs.~presentational \citep{mann1987}, and internal vs.~interpersonal \citep{hovy1995}.} This distinction is relevant for the data used in the current study, because the goals of RST-DT annotations and PDTB 2.0 annotations differ with respect to these functions. While annotators in RST-DT were instructed to annotate the writer's goal or intended effect of each segment of a text with respect to the neighbouring segments, PDTB annotators were asked to assess the relation between relational arguments, with a strong focus on the role of connectives (lexically-driven approach). 

The analysed empirical mappings provide support for the idea to annotate several levels of discourse relational arguments when appropriate. For instance,~a segment can be an example for something that was said in the other segment, but it may at the same time serve as evidence for a claim \cite[see also][]{carston1993,blakemore1997}.  Some of these patterns are systematic and go beyond the PDTB approach of allowing to annotate multiple labels for the same relation: PDTB annotators were not asked to systematically try to annotate all relations that hold. 
We conclude that future research should explore in more depth whether it is possible to devise an annotation procedure that allows researchers to identify both functions in a systematic way \cite[see also][]{crible2017}. 

Finally, regarding the annotation of explicit relations, we found that disagreements are often related to ambiguous connectives such as {\em as}, {\em but} and {\em while}. We analysed whether the annotations of ambiguous connectives agreed more with each other than can be expected based on the distribution of connectives in the data. We found that the PDTB and RST corpus annotations agreed well for relations marked by connectives that can mark very different types of relations (e.g., {\em while} can mark a causal or a temporal relation), but they disagreed often on annotation of connectives that mark similar types of relations (e.g., {\em but} can mark a contrastive or a concessive relation). We believe that these cases warrant further study in order to better understand why the disagreements occur, and what the implications should be (e.g., a less fine-grained distinction among discourse relations if the present distinction cannot be made reliably, or an improved operationalization of the annotation process in order to achieve more agreement?).


\paragraph{Future directions}
The mapped annotations will be made available online so that other researchers can profit from the aligned corpus. We see several possible directions of research for which this mapped data can be useful. First, for theoretical studies, the data can serve to further investigate the frameworks and the effects of their operationalizations on the annotations. Especially the cases which could not be aligned due to differences in structuring the discourse, and cases where different labels were chosen, are interesting from this viewpoint. Some of the mismatches that occur between the PDTB and RST-DT annotations are systematic; for example, certain causal labels in RST-DT are often annotated as additive labels in PDTB, and RST's \rel{Contrast} is often annotated in PDTB as \rel{Concession}.
The mapping reveals these patterns and can therefore function as a starting point for other experiments that investigate these systematic mismatches. 
%

Second, the mapping can prove to be useful for future annotations. The patterns of matches and mismatches that can be observed in the data can function as input for defining future annotation guidelines. The mapped data reveals which relation types may be particularly relevant for carrying several functions, and hence displayed less agreement between frameworks. The labels and definitions agreed upon across frameworks can be considered well-established, but for other types of relations, our mapping indicates that definitions may need to be refined in future efforts (for example, PDTB's and RST's \rel{Contrast} and \rel{Concession}, but also RST's \rel{Comparison} deserves more consideration). 
Our detailed results can also inform the ongoing discussion on identifying a set of labels for discourse relation annotation, which has been a long-lasting issue causing a lot of controversy in the literature.

Third, the mapped data can contribute towards automated discourse parsing efforts.
Discourse relation annotations have been used as training data in all recent efforts in automatic discourse relation classification. The classification of implicit discourse relations has received the bulk of the attention and work, given that classification of explicit relations was found to be relatively easy and accurate \cite{pitler2008}. Implicit discourse relation classification has recently also been the subject of two CoNNL shared tasks \cite{xue2015conll,xue2016conll}, with accuracies just over 40\% F-score on implicit relation sense labelling for an 11-way classification. Important questions to be considered in the light of the mapping results in this article relate to how these classification results can be interpreted in the light of the difficulty of the implicit relation classification task. How can we make sure that consistency is improved for training automatic discourse relation classifiers? Can and should we train classifiers separately for ideational vs.~intentional discourse relation levels? Should classifiers be evaluated by taking into account several possible labels for a relation, so that either the PDTB label or the corresponding RST label would be considered correct? Or should we weigh differently classification mismatch for categories that humans don't commonly replace for one another versus those that are more interchangeable?

Finally, we would like to emphasize that some of the methodological decisions for the present mapping are specific to the two exact frameworks we worked with, RST-DT and PDTB2.0. Other instances of RST-style annotation may treat nuclearity differently, in which case some of the assumptions of our alignment algorithm would not necessarily generalize to those annotations. Furthermore, PDTB-style annotation also differs between languages; different PDTB resources may use different relation inventories; nevertheless, we would expect that some of the fundamental observations we made (such as the effect of operationalization like the usage of implicit connectives during annotation) would transfer to those other PDTB-style resources. 


\bibliography{mybibfile.bib}

\begin{thebibliography}{45}
\expandafter\ifx\csname natexlab\endcsname\relax\def\natexlab#1{#1}\fi

\bibitem[{Asr and Demberg(2013)}]{asr2013}
Asr, Fatemeh~Torabi and Vera Demberg. 2013.
\newblock On the information conveyed by discourse markers.
\newblock In \emph{Proceedings of the Fourth Annual Workshop on Cognitive
  Modeling and Computational Linguistics}, pages 84--93.

\bibitem[{Benamara and Taboada(2015)}]{benamara2015}
Benamara, Farah and Maite Taboada. 2015.
\newblock Mapping different rhetorical relation annotations: A proposal.
\newblock In \emph{Proceedings of the Fourth Joint Conference on Lexical and
  Computational Semantics,* SEM}, pages 147--152.

\bibitem[{Blakemore(1997)}]{blakemore1997}
Blakemore, Diane. 1997.
\newblock Restatement and exemplification: A relevance theoretic reassessment
  of elaboration.
\newblock \emph{Pragmatics \& Cognition}, 5(1):1--19.

\bibitem[{Bunt and Prasad(2016)}]{bunt2016}
Bunt, Harry and Rashmi Prasad. 2016.
\newblock {ISO-DR-Core} ({ISO} 24617-8): Core concepts for the annotation of
  discourse relations.
\newblock In \emph{Proceedings 12th Joint ACL-ISO Workshop on Interoperable
  Semantic Annotation (ISA-12)}, pages 45--54.

\bibitem[{Carlson and Marcu(2001)}]{carlson2001}
Carlson, Lynn and Daniel Marcu. 2001.
\newblock \emph{Discourse tagging reference manual}.

\bibitem[{Carlson, Marcu, and Okurowski(2003)}]{carlson2003}
Carlson, Lynn, Daniel Marcu, and Mary~Ellen Okurowski. 2003.
\newblock Building a discourse-tagged corpus in the framework of {Rhetorical}
  {Structure} {Theory}.
\newblock In \emph{Current and new directions in discourse and dialogue}.
  Springer, pages 85--112.

\bibitem[{Carston(1993)}]{carston1993}
Carston, Robyn. 1993.
\newblock Conjunction, explanation and relevance.
\newblock \emph{Lingua}, 90(1-2):27--48.

\bibitem[{Chiarcos(2014)}]{chiarcos2014}
Chiarcos, Christian. 2014.
\newblock Towards interoperable discourse annotation. {Discourse} features in
  the {Ontologies} of {Linguistic} {Annotation}.
\newblock In \emph{Proceedings of the Ninth International Conference on
  Language Resources and Evaluation (LREC 2014)}, pages 4569--4577.

\bibitem[{Crible and Degand(2017)}]{crible2017}
Crible, Ludivine and Liesbeth Degand. 2017.
\newblock Reliability vs.~granularity in discourse annotation: What is the
  trade-off?
\newblock \emph{Corpus Linguistics and Linguistic Theory}.

\bibitem[{Hoek, Evers-Vermeul, and Sanders(2017)}]{hoek2017}
Hoek, Jet, Jacqueline Evers-Vermeul, and Ted~JM Sanders. 2017.
\newblock Segmenting discourse: Incorporating interpretation into segmentation?
\newblock \emph{Corpus Linguistics and Linguistic Theory}, Advance online
  publication.

\bibitem[{Hovy and Maier(1995)}]{hovy1995}
Hovy, Eduard~H and Elisabeth Maier. 1995.
\newblock Parsimonious or profligate: How many and which discourse structure
  relations.
\newblock \emph{Unpublished manuscript}.

\bibitem[{Jansen, Surdeanu, and Clark(2014)}]{jansen2014discourse}
Jansen, Peter, Mihai Surdeanu, and Peter Clark. 2014.
\newblock Discourse complements lexical semantics for non-factoid answer
  reranking.
\newblock In \emph{ACL (1)}, pages 977--986.

\bibitem[{Lee et~al.(2006)Lee, Prasad, Joshi, Dinesh, and Webber}]{lee2006}
Lee, Alan, Rashmi Prasad, Aravind Joshi, Nikhil Dinesh, and Bonnie Webber.
  2006.
\newblock Complexity of dependencies in discourse: Are dependencies in
  discourse more complex than in syntax?
\newblock In \emph{Proceedings of the 5th International Workshop on Treebanks
  and Linguistic Theories (TLT), Prague, Czech Republic}, pages 79--90.

\bibitem[{Mann and Thompson(1987)}]{mann1987}
Mann, William~C and Sandra~A Thompson. 1987.
\newblock \emph{Rhetorical {Structure} {Theory}: A theory of text
  organization}.
\newblock University of Southern California, Information Sciences Institute.

\bibitem[{Mann and Thompson(1988)}]{mann1988}
Mann, William~C and Sandra~A Thompson. 1988.
\newblock Rhetorical {Structure} {Theory}: Toward a functional theory of text
  organization.
\newblock \emph{Text-Interdisciplinary Journal for the Study of Discourse},
  8(3):243--281.

\bibitem[{Marcu(2000)}]{marcu2000}
Marcu, Daniel. 2000.
\newblock \emph{The theory and practice of discourse parsing and
  summarization}.
\newblock MIT press.

\bibitem[{Meyer and Popescu-Belis(2012)}]{meyer2012using}
Meyer, Thomas and Andrei Popescu-Belis. 2012.
\newblock Using sense-labeled discourse connectives for statistical machine
  translation.
\newblock In \emph{Proceedings of the Joint Workshop on Exploiting Synergies
  between Information Retrieval and Machine Translation (ESIRMT) and Hybrid
  Approaches to Machine Translation (HyTra)}, pages 129--138, Association for
  Computational Linguistics.

\bibitem[{Moore and Pollack(1992)}]{moore1992}
Moore, Johanna~D and Martha~E Pollack. 1992.
\newblock A problem for {RST}: The need for multi-level discourse analysis.
\newblock \emph{Computational Linguistics}, 18(4):537--544.

\bibitem[{Oza et~al.(2009)Oza, Prasad, Kolachina, Sharma, and Joshi}]{oza2009}
Oza, Umangi, Rashmi Prasad, Sudheer Kolachina, Dipti~Misra Sharma, and Aravind
  Joshi. 2009.
\newblock The {Hindi} discourse relation bank.
\newblock In \emph{Proceedings of the third Linguistic Annotation Workshop
  (LAW)}, pages 158--161, Association for Computational Linguistics.

\bibitem[{Pitler et~al.(2008)Pitler, Raghupathy, Mehta, Nenkova, Lee, and
  Joshi}]{pitler2008}
Pitler, Emily, Mridhula Raghupathy, Hena Mehta, Ani Nenkova, Alan Lee, and
  Aravind~K Joshi. 2008.
\newblock Easily identifiable discourse relations.
\newblock Technical report.

\bibitem[{Popescu-Belis(2016)}]{popescu2016manual}
Popescu-Belis, Andrei. 2016.
\newblock Manual and automatic labeling of discourse connectives for machine
  translation.
\newblock In \emph{TextLink--Structuring Discourse in Multilingual Europe
  Second Action Conference K{\'a}roli G{\'a}sp{\'a}r University of the Reformed
  Church in Hungary Budapest, 11--14 April, 2016}, page~16.

\bibitem[{Prasad et~al.(2008)Prasad, Dinesh, Lee, Miltsakaki, Robaldo, Joshi,
  and Webber}]{prasad2008}
Prasad, Rashmi, Nikhil Dinesh, Alan Lee, Eleni Miltsakaki, Livio Robaldo,
  Aravind~K. Joshi, and Bonnie Webber. 2008.
\newblock The {Penn} {Discourse} {TreeBank} 2.0.
\newblock In \emph{Proceedings of the International Conference on Language
  Resources and Evaluation (LREC)}, Citeseer.

\bibitem[{Prasad, Forbes-Riley, and Lee(2017)}]{prasad2017}
Prasad, Rashmi, Katherine Forbes-Riley, and Alan Lee. 2017.
\newblock Towards full text shallow discourse relation annotation: Experiments
  with cross-paragraph implicit relations in the {PDTB}.
\newblock In \emph{Proceedings of the 18th Annual SIGdial Meeting on Discourse
  and Dialogue}, pages 7--16.

\bibitem[{Prasad et~al.(2007)Prasad, Miltsakaki, Dinesh, Lee, Joshi, Robaldo,
  and Webber}]{prasad2007}
Prasad, Rashmi, Eleni Miltsakaki, Nikhil Dinesh, Alan Lee, Aravind~K. Joshi,
  Livio Robaldo, and Bonnie Webber. 2007.
\newblock \emph{The {Penn} {Discourse} {Treebank} 2.0 annotation manual}.

\bibitem[{Prasad, Webber, and Joshi(2014)}]{prasad2014reflections}
Prasad, Rashmi, Bonnie Webber, and Aravind Joshi. 2014.
\newblock Reflections on the {Penn} {Discourse} {Treebank}, comparable corpora,
  and complementary annotation.
\newblock \emph{Computational Linguistics}.

\bibitem[{Redeker(1990)}]{redeker1990}
Redeker, Gisela. 1990.
\newblock Ideational and pragmatic markers of discourse structure.
\newblock \emph{Journal of Pragmatics}, 14(3):367--381.

\bibitem[{Rehbein, Scholman, and Demberg(2016)}]{rehbein2016}
Rehbein, Ines, Merel C.~J. Scholman, and Vera Demberg. 2016.
\newblock Annotating discourse relations in spoken language: A comparison of
  the {PDTB} and {CCR} frameworks.
\newblock In \emph{Proceedings of the Tenth International Conference on
  Language Resources and Evaluation (LREC)}, European Language Resources
  Association (ELRA), Portoroz, Slovenia.

\bibitem[{Riloff and Wiebe(2003)}]{riloff2003}
Riloff, Ellen and Janyce Wiebe. 2003.
\newblock Learning extraction patterns for subjective expressions.
\newblock In \emph{Proceedings of the 2003 conference on Empirical Methods in
  Natural Language Processing}, pages 105--112, Association for Computational
  Linguistics.

\bibitem[{Robaldo and Miltsakaki(2014)}]{robaldo2014}
Robaldo, Livio and Eleni Miltsakaki. 2014.
\newblock Corpus-driven semantics of concession: Where do expectations come
  from?
\newblock \emph{Dialogue \& Discourse}, 5(1):1--36.

\bibitem[{Sanders et~al.(under review)Sanders, Demberg, Hoek, Scholman,
  Torabi~Asr, Zufferey, and Evers-Vermeul}]{sanderssubm}
Sanders, Ted J.~M., Vera Demberg, Jet Hoek, Merel C.~J. Scholman, Fatemeh
  Torabi~Asr, Sandrine Zufferey, and Jacqueline Evers-Vermeul. under review.
\newblock Unifying dimensions in discourse relations: How various annotation
  frameworks are related.
\newblock \emph{Corpus Linguistics and Linguistic Theory}.

\bibitem[{Sanders, Spooren, and Noordman(1992)}]{sanders1992}
Sanders, Ted J.~M., Wilbert P. M.~S. Spooren, and Leo G.~M. Noordman. 1992.
\newblock Toward a taxonomy of coherence relations.
\newblock \emph{Discourse Processes}, 15(1):1--35.

\bibitem[{Sanders et~al.(2016)Sanders, Demberg, Hoek, Scholman, Zufferey, and
  Evers-Vermuel}]{sanders2016}
Sanders, Ted~JM, Vera Demberg, Jet Hoek, Merel~CJ Scholman, Sandrine Zufferey,
  and Jacqueline Evers-Vermuel. 2016.
\newblock How can we relate various annotation schemes? {Unifying} dimensions
  in discourse relations.
\newblock In \emph{TextLink Second Action Conference}, pages 110--112.

\bibitem[{Scheffler and Stede(2016)}]{scheffler2016konvens}
Scheffler, Tatjana and Manfred Stede. 2016.
\newblock Mapping {PDTB}-style connective annotation to {RST}-style discourse
  annotation.
\newblock In \emph{Proceedings of the 13th Conference on Natural Language
  Processing (KONVENS 2016)}.

\bibitem[{Scholman and Demberg(2017)}]{scholmansubm}
Scholman, Merel C.~J. and Vera Demberg. 2017.
\newblock Examples and specifications that prove a point: Identifying
  elaborative and argumentative discourse relations.
\newblock \emph{Dialogue \& Discourse}, 8(2):56--83.

\bibitem[{Sharp et~al.(2015)Sharp, Jansen, Surdeanu, and
  Clark}]{sharp2015spinning}
Sharp, Rebecca, Peter Jansen, Mihai Surdeanu, and Peter Clark. 2015.
\newblock Spinning straw into gold: Using free text to train monolingual
  alignment models for non-factoid question answering.
\newblock In \emph{HLT-NAACL}, pages 231--237.

\bibitem[{Somasundaran et~al.(2009)Somasundaran, Namata, Wiebe, and
  Getoor}]{somasundaran2009supervised}
Somasundaran, Swapna, Galileo Namata, Janyce Wiebe, and Lise Getoor. 2009.
\newblock Supervised and unsupervised methods in employing discourse relations
  for improving opinion polarity classification.
\newblock In \emph{Proceedings of the 2009 Conference on Empirical Methods in
  Natural Language Processing: Volume 1-Volume 1}, pages 170--179, Association
  for Computational Linguistics.

\bibitem[{Stede(2008)}]{stede2008rst}
Stede, Manfred. 2008.
\newblock {RST} revisited: Disentangling nuclearity.
\newblock \emph{Subordination’versus’ Coordination’in Sentence and Text},
  pages 33--59.

\bibitem[{Stede and Neumann(2014{\natexlab{a}})}]{stede2014}
Stede, Manfred and Arne Neumann. 2014{\natexlab{a}}.
\newblock Potsdam {Commentary} {Corpus} 2.0: Annotation for discourse research.
\newblock In \emph{Proceedings of the Ninth International Conference on
  Language Resources and Evaluation (LREC 2014)}, pages 925--929.

\bibitem[{Stede and Neumann(2014{\natexlab{b}})}]{stede2014potsdam}
Stede, Manfred and Arne Neumann. 2014{\natexlab{b}}.
\newblock Potsdam {Commentary} {Corpus} 2.0: Annotation for discourse research.
\newblock In \emph{LREC}, pages 925--929.

\bibitem[{Webber et~al.(2016)Webber, Prasad, Lee, and
  Joshi}]{webber2016discourse}
Webber, Bonnie, Rashmi Prasad, Alan Lee, and Aravind Joshi. 2016.
\newblock A discourse-annotated corpus of conjoined {VPs}.
\newblock \emph{LAW X}, page~22.

\bibitem[{Xue et~al.(2015)Xue, Ng, Pradhan, Prasad, Bryant, and
  Rutherford}]{xue2015conll}
Xue, Nianwen, Hwee~Tou Ng, Sameer Pradhan, Rashmi Prasad, Christopher Bryant,
  and Attapol Rutherford. 2015.
\newblock The {CoNLL}-2015 shared task on shallow discourse parsing.
\newblock In \emph{CoNLL Shared Task}, pages 1--16.

\bibitem[{Xue et~al.(2016)Xue, Ng, Rutherford, Webber, Wang, and
  Wang}]{xue2016conll}
Xue, Nianwen, Hwee~Tou Ng, Attapol Rutherford, Bonnie Webber, Chuan Wang, and
  Hongmin Wang. 2016.
\newblock {CoNLL} 2016 shared task on multilingual shallow discourse parsing.
\newblock \emph{Proceedings of the CoNLL-16 shared task}, pages 1--19.

\bibitem[{Zhou et~al.(2011)Zhou, Li, Gao, Wei, and Wong}]{zhou2011unsupervised}
Zhou, Lanjun, Binyang Li, Wei Gao, Zhongyu Wei, and Kam-Fai Wong. 2011.
\newblock Unsupervised discovery of discourse relations for eliminating
  intra-sentence polarity ambiguities.
\newblock In \emph{Proceedings of the Conference on Empirical Methods in
  Natural Language Processing}, pages 162--171, Association for Computational
  Linguistics.

\bibitem[{Zirn et~al.(2011)Zirn, Niepert, Stuckenschmidt, and
  Strube}]{zirn2011fine}
Zirn, C{\"a}cilia, Mathias Niepert, Heiner Stuckenschmidt, and Michael Strube.
  2011.
\newblock Fine-grained sentiment analysis with structural features.
\newblock In \emph{IJCNLP}, pages 336--344.

\bibitem[{Zufferey and Degand(2013)}]{zufferey2013}
Zufferey, Sandrine and Liesbeth Degand. 2013.
\newblock Annotating the meaning of discourse connectives in multilingual
  corpora.
\newblock \emph{Corpus Linguistics and Linguistic Theory}, 1:1--24.

\end{thebibliography}
\newpage
\appendix

\section{Appendix A: PDTB 2.0 and RST-DT tagsets}
\label{appA}
\begin{figure}[h]
\caption{Hierarchy of relation senses in PDTB \cite{prasad2008}}
\centering
\includegraphics[scale = 0.45, clip=true, trim=0cm 10cm 0cm 0cm]{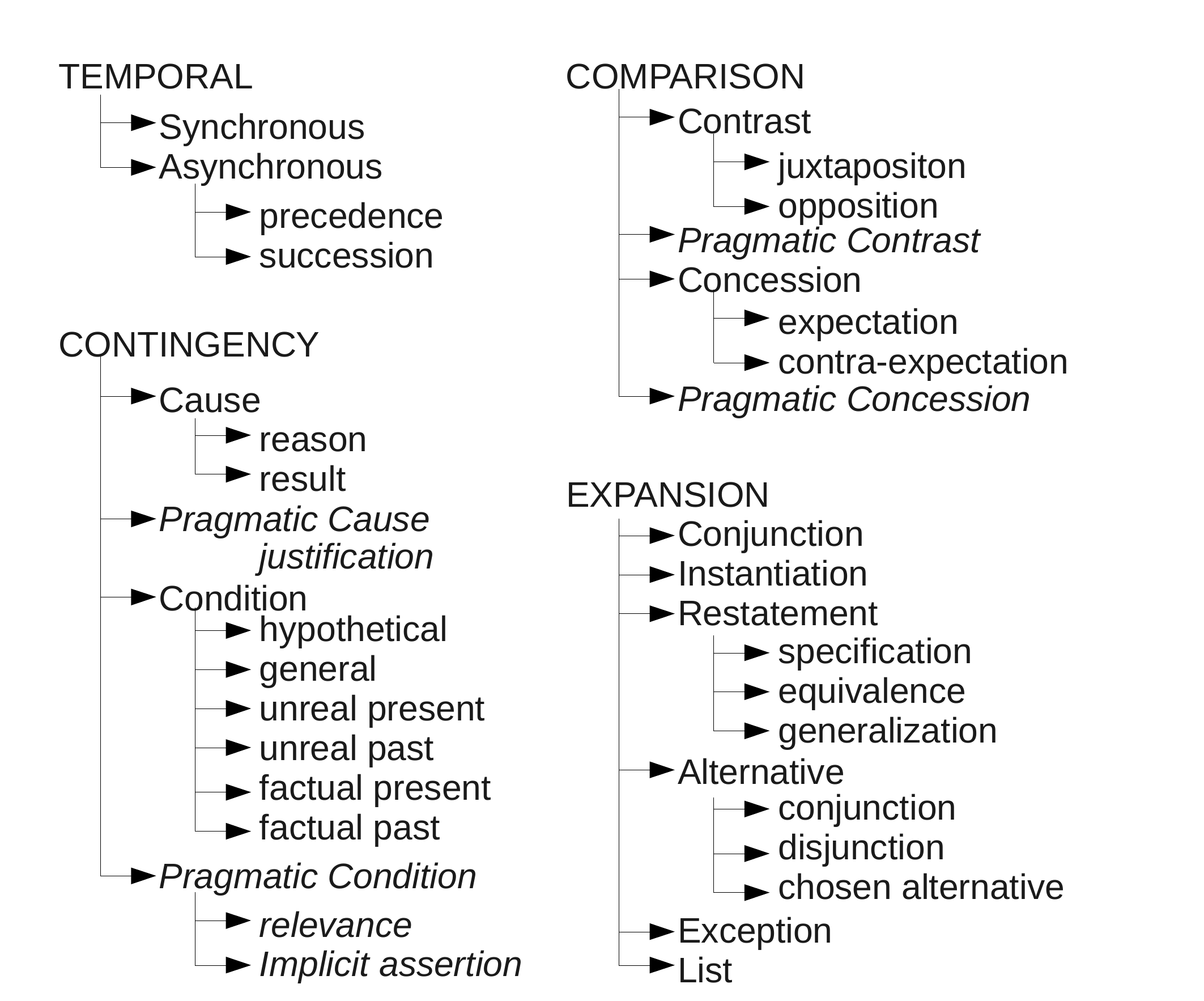}
\label{pdtbhierarchy}
\end{figure}

\begin{figure}[h]
\caption{Tagset of relation senses in RST-DT \citep{carlson2001}} 
\centering
\includegraphics[width=\textwidth]{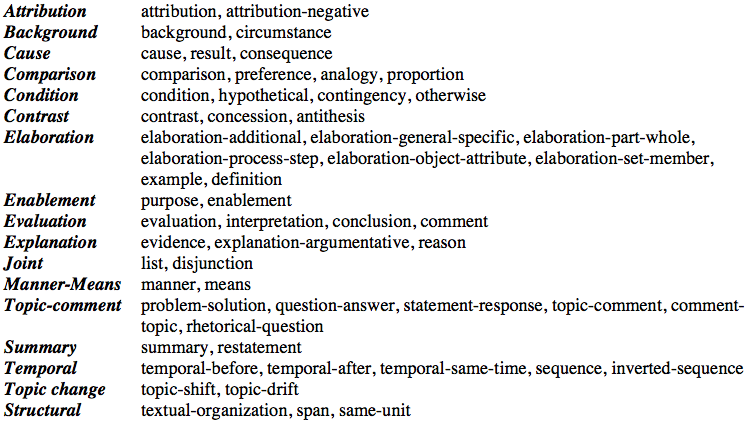}
\label{fig:rsttagset}
\end{figure}

\newpage

\section{Appendix B: ISO tagset and Unifying Dimensions definitions}
\label{appB}

\begin{figure}[h]
\caption{Tagset of relation senses in the ISO proposal \citep{bunt2016}} 
\centering
\includegraphics[width=0.98\textwidth]{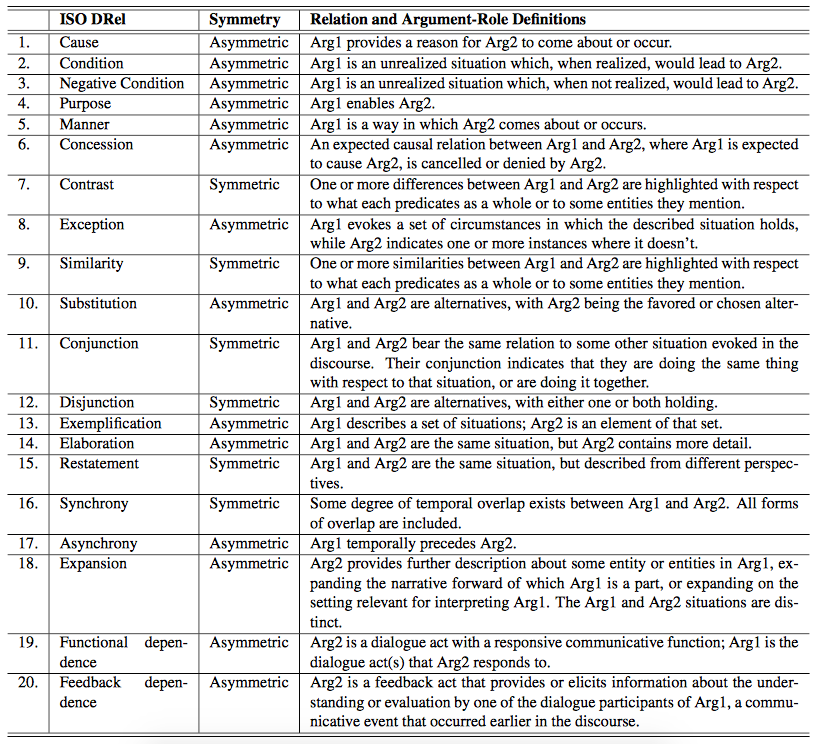}
\label{fig:isotagset}
\end{figure}

\newpage

\begin{figure}[h]
\caption{The unifying dimensions and features, and their values \cite{sanderssubm}} 
\centering
\includegraphics[width=1.03\textwidth]{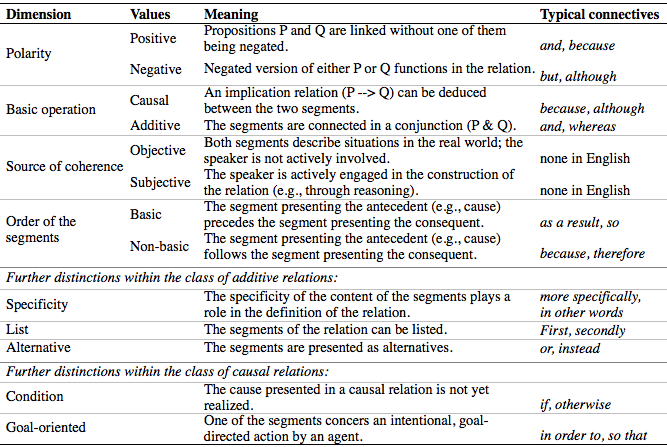}
\label{fig:unidimdef}
\end{figure}

\newpage

\section{Appendix C: ISO-based and Unifying Dimension-based mappings between RST-DT and PDTB 2.0 labels}
\label{appC}

\footnotesize{Note:  Labels that were excluded by at least two frameworks were not included in this table. These labels are RST-DT's \rel{Question-answer}, \rel{Statement-response}, \rel{Topic-comment}, \rel{Comment-topic}, \rel{Rhetorical-question}, \rel{Topic-shift}, \rel{Topic-drift}, \rel{Attribution}.} General labels for the types \rel{Condition}, \rel{Contrast}, and \rel{Expansion} were included as subtypes because the three proposals mapped to these more general labels.

\begin{table}[b]
\caption{Proposed mappings between RST-DT and PDTB 2.0 \rel{Temporal}, \rel{Contingency}, and \rel{Comparison} labels (\rel{Expansion} labels on next page). Cells with an `o' indicate proposed correspondence according to the OLiA, cells with a `u' according to the Unifying Dimensions proposal, and cells with `i' according to the ISO proposal.}
\scriptsize
\label{tab:theoreticMapping}
\begin{tabular}{l||@{\hskip3pt}c@{\hskip3pt}@{\hskip3pt}c@{\hskip3pt}@{\hskip3pt}c@{\hskip3pt}||@{\hskip3pt}c@{\hskip3pt}@{\hskip3pt}c@{\hskip3pt}@{\hskip3pt}c@{\hskip3pt}@{\hskip3pt}c@{\hskip3pt}@{\hskip3pt}c@{\hskip3pt}@{\hskip3pt}c@{\hskip3pt}||@{\hskip3pt}c@{\hskip3pt}@{\hskip3pt}c@{\hskip3pt}@{\hskip3pt}c@{\hskip3pt}@{\hskip3pt}c@{\hskip3pt}@{\hskip3pt}c@{\hskip3pt}@{\hskip3pt}c@{\hskip3pt}@{\hskip3pt}c@{\hskip3pt}|}

PDTB $\rightarrow$        & \multicolumn{3}{c||@{\hskip3pt}}{Temporal}                            & \multicolumn{6}{c||@{\hskip3pt}}{Contingency}                                                                             & \multicolumn{7}{c|}{Comparison}                                            \\

                                       & \multicolumn{1}{c|}{} & \multicolumn{2}{c||@{\hskip3pt}}{Asynch.}                            & \multicolumn{2}{c|@{\hskip3pt}}{Cause}  & \multicolumn{1}{c|@{\hskip3pt}}{Pragm.} & \multicolumn{1}{c|@{\hskip3pt}}{}    & \multicolumn{2}{c||@{\hskip3pt}}{Prag.}    & \multicolumn{3}{c|@{\hskip3pt}}{Contrast}      & \multicolumn{1}{c|@{\hskip3pt}}{}        & \multicolumn{2}{c|@{\hskip3pt}}{Concess.} & \multicolumn{1}{c|}{}                                 \\
                                       & \multicolumn{1}{c|}{} & \multicolumn{2}{c||@{\hskip3pt}}{}                            & \multicolumn{2}{c|@{\hskip3pt}}{}  & \multicolumn{1}{c|@{\hskip3pt}}{cause} & \multicolumn{1}{c|@{\hskip3pt}}{}    & \multicolumn{2}{c||@{\hskip3pt}}{cond.}    & \multicolumn{3}{c|@{\hskip3pt}}{}      & \multicolumn{1}{c|@{\hskip3pt}}{}        & \multicolumn{2}{c|@{\hskip3pt}}{} & \multicolumn{1}{c|}{}                                 \\

RST-DT  $\downarrow$       & \multicolumn{1}{c|}{\sid{Synchronous}} & \sid{Precedence} & \sid{Succession} & \sid{Reason} & \multicolumn{1}{c|@{\hskip3pt}}{\sid{Result}} & \multicolumn{1}{c|@{\hskip3pt}}{\sid{Justification}} & \multicolumn{1}{c|@{\hskip3pt}}{\sid{Condition}} & \sid{Relevance} & \sid{Impl. assert.} & \sid{Contrast} & \sid{Juxtapos.} & \multicolumn{1}{c|@{\hskip3pt}}{\sid{Opposition}} & \multicolumn{1}{c|@{\hskip3pt}}{\sid{Pragm. contr.}}     & \sid{Expectation} &\multicolumn{1}{c|@{\hskip3pt}}{ \sid{Contra-exp.}} & \sid{Pragm. conc.} \\ \hline

Background         &                   & u                & u                &              &              &                     &                 &                 &                     &                &                 &                  &                  & & &    \\
Circumstance       & u, i                 & u                & u                &              &              &                     &                 &                 &                     &                &                 &                  &                 & & &     \\ \hline
Cause                  &                   &                  &                  & o, u, i        & o, u, i         & i                   &                 &                 &                     &                &                 &                  &                 & & &     \\
Cause-result          &                   &                  &                  & o, u, i         & o, u, i         & i                   &                 &                 &                     &                &                 &                  &                & & &      \\
Result                    &                   &                  &                  & o, u, i         & o, u, i         & i                   &                 &                 &                     &                &                 &                  &                 & & &     \\
Consequence        &                   &                  &                  & o, u, i         & o, u, i         & i                   &                 &                 &                     &                &                 &                  &                  & & &    \\ \hline
Comparison         &                   &                  &                  &              &              &                     &                 &                 &                     & i              & i               & i                &                  & & &    \\
Preference         &                   &                  &                  &              &              &                     &                 &                 &                     & u              & u               & u                & u              & i & i &      \\
Analogy            &                   &                  &                  &              &              &                     &                 &                 &                     &                &                 &                  &                  & & &    \\
Proportion         &                   &                  &                  &              &              &                     & u               & u               & u                   &                &                 &                  &                  & & &    \\ \hline
Condition          &                   &                  &                  &              &              &                     & o, u, i            & o, u               & o, u                   &                &                 &                  &                 & & &     \\
Hypothetical       &                   &                  &                  &              &              &                     & o, u, i            & u               & u                   &                &                 &                  &                  & & &    \\
Contingency        &                   &                  &                  &              &              &                     & o, u, i            &                 &                     &                &                 &                  &                   & & &   \\
Otherwise          &                   &                  &                  &              &              &                     & u, i            &                 &                     &                &                 &                  &                  & & &    \\ \hline
Contrast           &                   &                  &                  &              &              &                     &                 &                 &                     & o, u, i           & o, u               & o, u                & o, u                & & &   \\
Concession         &                   &                  &                  &              &              &                     &                 &                 &                     &                &                 &                  &                  & o, u, i  & o, u, i  & o, u   \\
Antithesis         &                   &                  &                  &              &              &                     &                 &                 &                     & o, u              & o, u               & o, u                & o, u                 & u, i  & u, i  & u   \\ \hline
El-additional     &                   &                  &                  &              &              &                     &                 &                 &                     &                &                 &                  &                  & & &    \\
El-gen.-spec.     &                   &                  &                  &              &              &                     &                 &                 &                     &                &                 &                  &                   & & &   \\
El-part-whole     &                   &                  &                  &              &              &                     &                 &                 &                     &                &                 &                  &                   & & &   \\
El-proc.-step   &                   &                  &                  &              &              &                     &                 &                 &                     &                &                 &                  &                   & & &   \\
El-object-attr.   &                   &                  &                  &              &              &                     &                 &                 &                     &                &                 &                  &                   & & &   \\
El-set-mem.     &                   &                  &                  &              &              &                     &                 &                 &                     &                &                 &                  &                   & & &   \\
Example            &                   &                  &                  &              &              &                     &                 &                 &                     &                &                 &                  &                   & & &   \\
Definition         &                   &                  &                  &              &              &                     &                 &                 &                     &                &                 &                  &                   & & &   \\ \hline
Purpose            &                   &                  &                  &    o          & o, u, i         &                     &                 &                 &                     &                &                 &                  &                 & & &     \\
Enablement         &                   &                  &                  &              & u, i            &                     &                 &   o              &   o                  &                &                 &                  &                   & & &   \\ \hline
Evaluation         &                   &                  &                  & u            &              & u                   &                 &                 &                     &                &                 &                  &                   & & &   \\
Conclusion         &                   &                  &                  & o, u            &     o         & o, u                   &     o            &         o        &         o            &                &                 &                  &                   & & &   \\ 
Comment            &                   &                  &                  &              &              &                     &                 &                 &                     &                &                 &                  &                  & & &    \\ \hline
Evidence           &                   &                  &                  & u, i         & u, i         & o, u, i                &                 &                 &                     &                &                 &                  &                   & & &   \\
Expl.-argum.    &                   &                  &                  & o, u, i         & o, u, i         & i                   &                 &                 &                     &                &                 &                  &                 & & &     \\
Reason             &                   &                  &                  & o, u, i         & o, u, i         & i                   &                 &                 &                     &                &                 &                  &                 & & &     \\ \hline
List               &                   &                  &                  &              &              &                     &                 &                 &                     &                &                 &                  &                  & & &    \\
Disjunction        &                   &                  &                  &              &              &                     &                 &                 &                     &                &                 &                  &                   & & &   \\ \hline
Summary            &                   &                  &                  &              &              &                     &                 &                 &                     &                &                 &                  &                & & &      \\
Restatement        &                   &                  &                  &              &              &                     &                 &                 &                     &                &                 &                  &                  & & &    \\ \hline
Temp.-before       &                   & o, u, i             &  i             &              &              &                     &                 &                 &                     &                &                 &                  &                 & & &     \\
Temp.-after        &                   & i             & o, u, i             &              &              &                     &                 &                 &                     &                &                 &                  &                & & &      \\
T.-same-time     & o, u, i              &                  &                  &              &              &                     &                 &                 &                     &                &                 &                  &                & & &      \\
Sequence           &                   & o, u, i             & u, i             &              &              &                     &                 &                 &                     &                &                 &                  &                & & &      \\
Inverted-seq.      &                   & u, i             & o, u, i             &              &              &                     &                 &                 &                     &                &                 &                  &                & & &      \\ \hline
Means              &                   &                  &                  & u            & u            &                     &                 &     o            &      o               &                &                 &                  &               & & &       \\ \hline
Problem-sol.     &                   &                  &                  & u            & u            &                     &                 &  o               &      o               &                &                 &                  &                & & &      \\ \hline
Not mapped         &                   &                  &                  &              &              &                     &                 & i               & i                   &                &                 &                  & i             & & & i      \\ \hline
\end{tabular}
\end{table}

\newpage

\begin{table}[h]
\caption{Proposed mappings between RST-DT and PDTB 2.0 \rel{Expansion} labels, including a general {\sc Expansion} category. Cells with an `o' indicate proposed correspondence according to the OLiA, cells with a `u' according to the Unifying Dimensions proposal, and cells with `i' according to the ISO proposal.}
\scriptsize
\begin{tabular}{l||@{\hskip3pt}c@{\hskip3pt}@{\hskip3pt}c@{\hskip3pt}@{\hskip3pt}c@{\hskip3pt}@{\hskip3pt}c@{\hskip3pt}@{\hskip3pt}c@{\hskip3pt}@{\hskip3pt}c@{\hskip3pt}@{\hskip3pt}c@{\hskip3pt}@{\hskip3pt}c@{\hskip3pt}@{\hskip3pt}c@{\hskip3pt}@{\hskip3pt}c@{\hskip3pt}@{\hskip3pt}c@{\hskip3pt}||@{\hskip3pt}c@{\hskip3pt}||@{\hskip3pt}c@{\hskip3pt}|}

PDTB  $\rightarrow$  & \multicolumn{11}{c||@{\hskip3pt}}{Expansion}     &        &      \\
                                 & \multicolumn{1}{c|@{\hskip3pt}}{}  & \multicolumn{1}{c|@{\hskip3pt}}{} & \multicolumn{1}{c|@{\hskip3pt}}{} & \multicolumn{3}{c|@{\hskip3pt}}{Restatement} &  \multicolumn{3}{c|@{\hskip3pt}}{Alternative} & \multicolumn{1}{c|@{\hskip3pt}}{}    &        &    &  \\

RST-DT  $\downarrow$  & \multicolumn{1}{c|@{\hskip3pt}}{\sid{Expansion}} & \multicolumn{1}{c|@{\hskip3pt}}{\sid{Conjunction}} & \multicolumn{1}{c|@{\hskip3pt}}{\sid{Instantiation}} & \sid{Specification} & \sid{Equivalence} & \multicolumn{1}{c|@{\hskip3pt}}{\sid{Generaliz.}} & \sid{Conjunctive} & \sid{Disjunctive} & \multicolumn{1}{c|@{\hskip3pt}}{\sid{Chosen alt.}} & \multicolumn{1}{c|@{\hskip3pt}}{\sid{Exception}} & \sid{List} & \sid{EntRel} & \sid{Not mapped} \\ \hline
Background          &     i         & u                 &                     &                     &                   &                  &                   &                   &                   &                 &            &      &   o     \\
Circumstance        &              & u                 &                     &                     &                   &                  &                   &                   &                   &                 &            &        &   o  \\ \hline
Cause                   &          &                   &                     &                     &                   &                  &                   &                   &                   &                 &            &        &      \\
Cause-result       &           &                     &                     &                   &                  &                   &                   &                   &                 &            &       &   &    \\
Result                  &       &                   &                     &                     &                   &                  &                   &                   &                   &                 &            &        &      \\
Consequence       &               &                   &                     &                     &                   &                  &                   &                   &                   &                 &            &         &     \\ \hline
Comparison          &              & u                 &                     &                     &                   &                  &                   &                   &                   &                 &            &        & o     \\
Preference         &   &                   &                     &                     &                   &                  &                   &                   & u                 & u               &            &        &   o   \\
Analogy               &          & u, i              &                     &                     &                   &                  &                   &                   &                   &                 &            &           &  o \\
Proportion           &   & u, i              &                     &                     &                   &                  &                   &                   &                   &                 &            &         &  o   \\ \hline
Condition             &    &                   &                     &                     &                   &                  &                   &                   &                   &                 &            &        &      \\
Hypothetical         &          &                   &                     &                     &                   &                  &                   &                   &                   &                 &            &         &     \\
Contingency       &         &                   &                     &                     &                   &                  &                   &                   &                   &                 &            &         &     \\
Otherwise        &     &                   &                     &                     &                   &                  &     o              &     o              &      o             &                 &            &         &     \\ \hline
Contrast        &    &                   &                     &                     &                   &                  &                   &                   &                   & u               &            &          &    \\
Concession      &     &                   &                     &                     &                   &                  &                   &                   &                   &                 &            &        &      \\
Antithesis     &   &                   &                     &                     &                   &                  &                   &                   & i                 & u               &            &          &    \\ \hline
El-additional    & o  & u, i              &                     &                     &                   &                  &                   &                   &                   &                 &            & i        &    \\
El-gen.-spec.   &  &                   &                     & o, u, i                &                   & o, u, i             &                   &                   &                   &                 &            &          &    \\
El-part-whole  & o &                   &                     & u, i                &                   & u, i             &                   &                   &                   &                 &            &          &    \\
El-proc.-step   &o  &                   &                     & u, i                &                   & u, i             &                   &                   &                   &                 &            &         &     \\
El-object-attr.       & o  &                   &                     & u                   &                   & u                &                   &                   &                   &                 &            & i       &     \\
El-set-mem.     &   o   &                   & u, i                &                     &                   &                  &                   &                   &                   &                 &            &        &      \\
Example           &     &                   & o, u, i                &                     &                   &                  &                   &                   &                   &                 &            &           &   \\
Definition         & i  &                   &                     & u                   &                   & u                &                   &                   &                   &                 &            &           & o, i  \\ \hline
Purpose           &    &                   &                     &                     &                   &                  &                   &                   &                   &                 &            &          &    \\
Enablement    &    &                   &                     &                     &                   &                  &                   &                   &                   &                 &            &         &     \\ \hline
Evaluation     &  i    &                   &                     & u                   &                   & u                &                   &                   &                   &                 &            &          &   o \\
Conclusion      &    &                   &                     & i                   &                   & i                &                   &                   &                   &                 &            &         &     \\
Comment       & i    & u                 &                     & u                   &                   & u                &                   &                   &                   &                 &            &         &  o   \\ \hline
Evidence       &   &                   &                     &                     &                   &                  &                   &                   &                   &                 &            &          &    \\
Expl.-argum.  &  &                   &                     &                     &                   &                  &                   &                   &                   &                 &            &         &     \\
Reason      &     &                   &                     &                     &                   &                  &                   &                   &                   &                 &            &         &     \\ \hline
List            &   &                   &                     &                     &                   &                  &                   &                   &                   &                 & o, u, i       &        &      \\
Disjunction   &    &                   &                     &                     &                   &                  & o, u, i              & o, u, i              & o, u                 &                 &            &          &    \\ \hline
Summary     &       &                   &                     & o, u, i                &   o                & o, u, i             &                   &                   &                   &                 &            &          &    \\
Restatement   &       &                   &                     &                     & o, u, i              &                  &                   &                   &                   &                 &            &        &      \\ \hline
Temp.-before  &    &                   &                     &                     &                   &                  &                   &                   &                   &                 &            &         &     \\
Temp.-after   &   &                   &                     &                     &                   &                  &                   &                   &                   &                 &            &         &     \\
T.-same-time    &    &                   &                     &                     &                   &                  &                   &                   &                   &                 &            &          &    \\
Sequence       &    &                   &                     &                     &                   &                  &                   &                   &                   &                 &            &        &      \\
Inverted-seq.  &  &                   &                     &                     &                   &                  &                   &                   &                   &                 &            &          &    \\ \hline
Means            &  &                   &                     &                     &                   &                  &                   &                   &                   &                 &            &          &  i  \\ \hline
Problem-sol.    & i  &                   &                     &                     &                   &                  &                   &                   &                   &                 &            &          &    \\ \hline
Not mapped    &      &                   &                     &                     &                   &                  &                   &                   &                   & o, i               &            & u     &    \\   \hline
\end{tabular}
\end{table}

\end{document}